%% file: main.tex
\documentclass[10pt,twocolumn,letterpaper]{article}

\usepackage{cvpr}              
\input{preamble}
\definecolor{cvprblue}{rgb}{0.21,0.49,0.74}
\usepackage[pagebackref,breaklinks,colorlinks,allcolors=cvprblue]{hyperref}

\usepackage{tabularx}
\usepackage{multirow}
\usepackage{appendix}
\usepackage{graphicx}   
\usepackage{array}      
\usepackage{adjustbox} 
\usepackage[linesnumbered,ruled,vlined]{algorithm2e}
\usepackage{amsmath}
\usepackage{booktabs}
\usepackage{soul, color, xcolor}
\newcolumntype{Y}{>{\centering\arraybackslash}X} 

\definecolor{bestred}{RGB}{255, 200, 190}
\definecolor{secondpurple}{RGB}{203, 203, 254}

\makeatletter
\newcommand{\removelatexerror}{\let\@latex@error\@gobble}
\makeatother

\def\paperID{16547} 
\def\confName{CVPR}
\def\confYear{2026}

\title{JointTuner: Appearance-Motion Adaptive Joint Training for Customized Video Generation}

\author{
    Fangda Chen$^{1}$, Shanshan Zhao$^{2}$, Chuanfu Xu$^{1}$, Long Lan$^{1}$\thanks{Corresponding author} \\
    $^{1}$College of Computer Science and Technology, National University of Defense Technology \\
    $^{2}$Alibaba International Digital Commerce \\
    {\tt\small \{fdchen.nudt, xuchuanfu, long.lan\}@nudt.edu.cn} \\
    {\tt\small sshan.zhao00@gmail.com}
}

\begin{document}
\twocolumn[{
\maketitle
\vspace{-4em}
\begin{center}
    \centering
    \includegraphics[width=\linewidth]{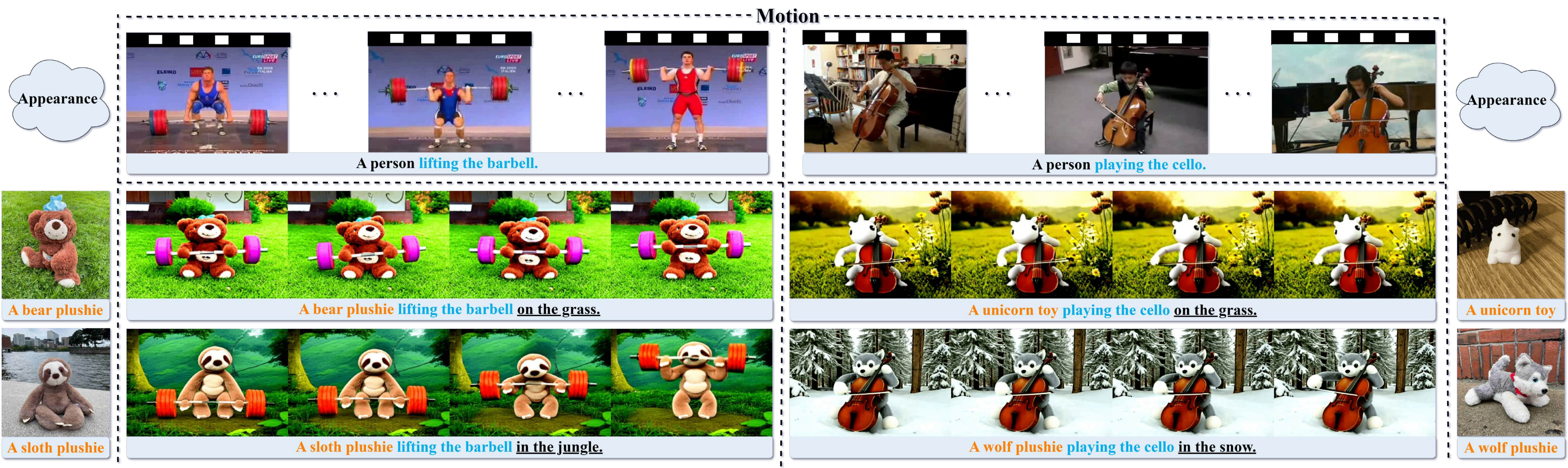}
    \vspace{-2em}
   \captionof{figure}{\textbf{Results of customized video generation using JointTuner.} Given paired appearance and motion inputs, it produces videos that reflect both the desired subject appearance and motion patterns through adaptive joint training.}
  \label{fig:teaser}
  \vspace{-0.5em}
\end{center}
}]

\input{0_abstract}    
\input{1_intro}
\input{2_relate}

\input{3_method}
\input{4_exp}
\input{5_conclusion}
{
    \clearpage
    \small
    \bibliographystyle{ieeenat_fullname}
    \bibliography{main}
}
\input{X_suppl}

\end{document}

%% file: 0_abstract.tex
\begin{abstract}

\vspace{-1em}
Recent advancements in customized video generation have led to significant improvements in the simultaneous adaptation of appearance and motion. 
Typically, decoupling the appearance and motion training, prior methods 
often introduce \textit{concept interference}, resulting in inaccurate rendering of appearance features or motion patterns. In addition, these methods often suffer from \textit{appearance contamination}, in which background and foreground elements from reference videos distort the customized video. 
This paper aims to alleviate these issues by proposing \textbf{JointTuner}. The core motivation of our JointTuner is to enable joint optimization of both appearance and motion components, upon which two key innovations are developed, i.e., \textbf{Gated Low-Rank Adaptation (GLoRA)} and \textbf{Appearance-independent Temporal Loss (AiT Loss)}. Specifically, GLoRA uses a context-aware activation layer, analogous to a gating regulator, to dynamically steer LoRA modules toward learning either appearance or motion while maintaining spatio-temporal consistency. Moreover, with the finding that channel-temporal shift noise suppresses appearance-related low-frequencies while enhancing motion-related high-frequencies, we designed the AiT Loss. This loss adds the same shift to the diffusion model’s predicted noise during fine-tuning, forcing the model to prioritize learning motion patterns.
JointTuner's architecture-agnostic design supports both UNet (e.g., ZeroScope) and Diffusion Transformer (e.g., CogVideoX) backbones, ensuring its customization capabilities scale with the evolution of foundational video models.
Furthermore, we present a systematic evaluation framework for appearance-motion combined customization, covering 90 combinations evaluated along four critical dimensions: semantic alignment, motion dynamism, temporal consistency, and perceptual quality. Our \href{https://fdchen24.github.io/JointTuner-Website}{project homepage} is available online.
\end{abstract}

%% file: 1_intro.tex
\section{Introduction}
\label{sec:intro}

\begin{figure*}[t]
    \centering
    \includegraphics[width=\linewidth]{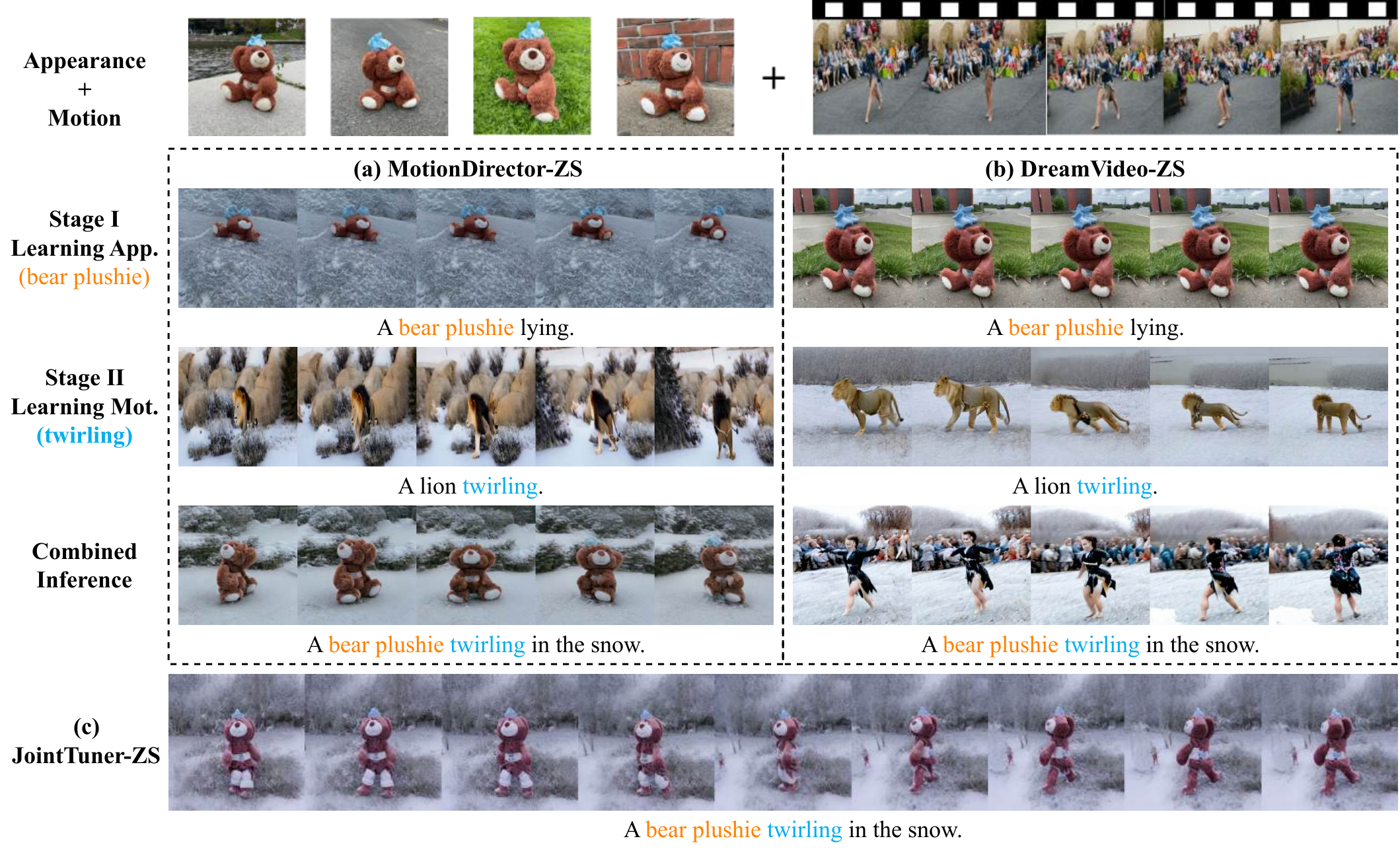}
    \vspace{-2em}
   \caption{Illustration of a failure case from advanced customized video generation methods. (a) and (b) show videos generated by MotionDirector-ZS~\cite{motiondirector} and DreamVideo-ZS~\cite{dreamvideo} across three stages: appearance learning (bear plushie), motion learning (twirling), and combined inference. (c) presents results from JointTuner-ZS, which improves inference by jointly learning both appearance and motion. Note that ``-ZS'' denotes methods based on ZeroScope~\cite{zeroscope}.}
  \label{fig:limitaion_sample}
  \vspace{-2em}
\end{figure*}

Recent advances in text-to-video generation have defined three core tasks in customized video generation (CVG): appearance customization using images to preserve subject identity~\cite{customvideo, dynvideo, customcrafter, tweediemix}; motion customization by transferring motion from reference videos~\cite{tuneavideo, motionshop, separatemotion}; and appearance-motion combined customization, which simultaneously controls both aspects~\cite{motiondirector, dreamvideo, motrans, customizeavideo, anyv2v, videoswap}. Appearance customization typically focuses on static attributes, while motion customization emphasizes dynamic behavior. For complex motions or fine-grained appearances beyond textual description, combining images for identity and videos for motion enables more precise results. In this work, we focus on this setting to customize both appearance and motion, aiming to enhance overall generation quality.

Existing appearance-motion combined customization methods typically adopt stage-wise training, separately optimizing appearance and motion components~\cite{motiondirector, dreamvideo}. Such a decoupled process often leads to \textit{concept interference}: as shown in Fig.~\ref{fig:limitaion_sample}(a), models struggle to generate desired motion due to poor spatial-temporal fusion. Another issue is \textit{appearance contamination}, where unwanted elements from reference videos leak into the output (Fig.~\ref{fig:limitaion_sample}(b)). Recently, Diffusion Transformer (DiT)-based text-to-video models~\cite{2022cogvideo, wan} have enabled higher-quality and longer video generation. Yet, simply inserting learnable parameters into DiT, training it stage-wise, and fusing the modules for inference still leads to \textit{concept interference}~\cite{dualreal}.

\begin{figure*}[t]
    \centering
    \includegraphics[width=\linewidth]{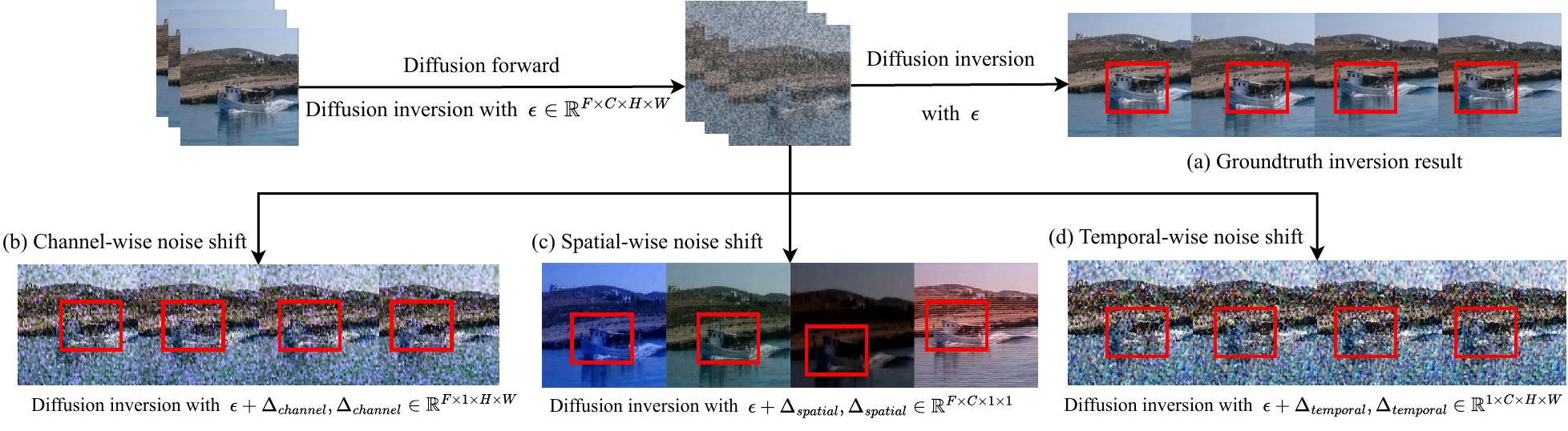}
    \vspace{-2em}
   \caption{\textbf{Illustration of the impact of noise shift on diffusion inversion in latent space with CogVideoX~\cite{2022cogvideo}.} Starting from a clean video, random noise $\epsilon \in \mathbb{R}^{F \times C \times H \times W}$ is added during the diffusion forward process, followed by diffusion inversion to recover the original video. Red bounding boxes highlight regions of interest.}
  \label{fig:different_shift_visual}
  \vspace{-1em}
\end{figure*}

\begin{figure*}[t]
    \centering
    \includegraphics[width=\linewidth]{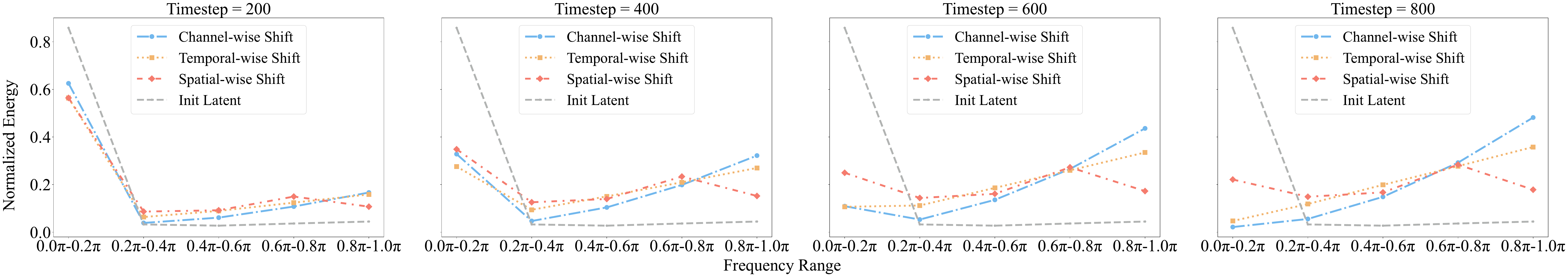}
    \vspace{-2em}
   \caption{\textbf{Frequency distribution of latent and shifted latent signals across shift types and time steps.} These results show normalized energy for three shift types: Channel-wise, Spatial-wise, and Temporal-wise at time steps 200, 400, 600, and 800.}
  \label{fig:different_shift_result}
  \vspace{-2em}
\end{figure*}

To overcome these challenges, we introduce JointTuner, an adaptive joint training framework designed to concurrently learn both appearance and motion. Our approach introduces two key innovations:
\textbf{(1) Gated Low-Rank Adaptation (GLoRA)}: To resolve conflicts between appearance and motion, we introduce GLoRA, an architecture that dynamically adjusts the activation of modality-specific expert modules. This is achieved through a gating regulator that modulates each LoRA based on the input modality. The gating regulator learns to activate the LoRAs focused on either appearance or motion during training, enabling dynamic composition of specialized modules at inference, suppressing concept interference, and allowing coherent joint modeling.
\textbf{(2) Appearance-independent Temporal Loss (AiT Loss)}: As shown in Fig.~\ref{fig:different_shift_visual}, channel- and temporal-wise shifts disrupt appearance yet preserve motion, while spatial shifts mainly change color tones. Spectral analysis (Fig.~\ref{fig:different_shift_result}) confirms that these shifts suppress low-frequency appearance signals and enhance high-frequency motion components~\cite{freelong, riflex}. Based on this, AiT Loss injects channel–temporal shift noise during training, guiding the model to learn motion patterns independent of appearance and suppress contamination from the reference video’s appearance. Consequently, as shown in Fig.~\ref{fig:limitaion_sample}(c), our method produces more complete motion and significantly reduces interference from reference video appearance.

While most existing appearance-motion customization methods~\cite{motiondirector, customttt, dualreal} rely on curated datasets and appearance-centric metrics like CLIP-Image or Subject Consistency, they often neglect motion evaluation due to the lack of a unified benchmark. To bridge this gap and enable balanced assessment, we introduce a comprehensive benchmark comprising 90 subject-motion combinations from public datasets~\cite{dreambooth, customdiffusion, motiondirector, davis2016, ucf101}. Our accompanying evaluation framework includes 10 metrics across four dimensions, semantic alignment, motion dynamism, temporal consistency, and perceptual quality, along with three composite scores, establishing a standardized protocol for appearance-motion combined customization.

Our contributions are as follows:

\begin{itemize}
\item We propose \textbf{JointTuner}, an adaptive joint training framework for appearance-motion combined customization, compatible with both UNet- and DiT-based architectures.

\item We introduce \textbf{GLoRA}, a gating adaptation mechanism that reduces concept interference and enables seamless integration of appearance and motion representations.

\item We find that injecting shift noise along channel and temporal dimensions suppresses appearance-related low-frequency components while enhancing motion-related high-frequency ones, leading to the design of \textbf{AiT Loss} to reduce appearance contamination.

\item We establish a comprehensive evaluation framework for appearance-motion combined customization, featuring diverse subject-motion combinations and metrics assessing semantic alignment, motion fidelity, temporal coherence, and perceptual quality.
\end{itemize}

%% file: 2_relate.tex
\section{Related Work}
\label{sec:relatedwork}

\begin{figure*}[t]
    \centering
    \includegraphics[width=\linewidth]{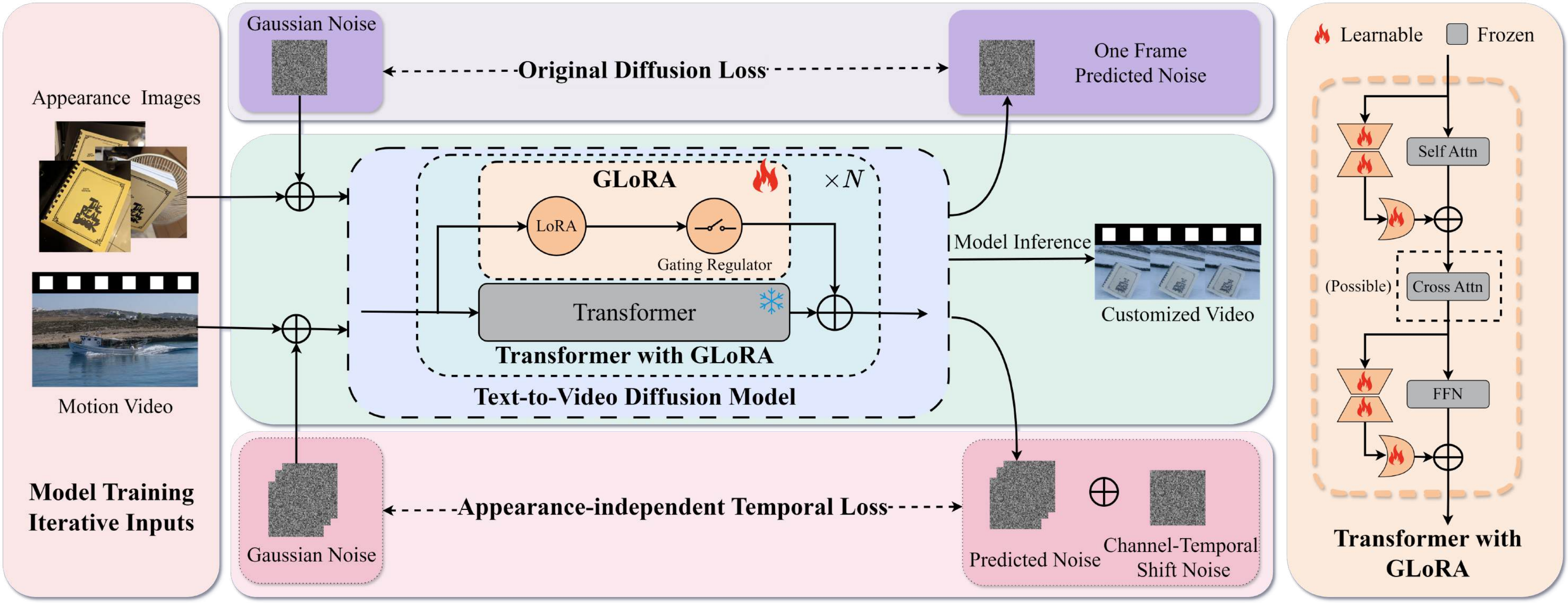}
    \vspace{-2em}
    \caption{\textbf{Architecture of JointTuner, an adaptive joint training framework with two main steps}: (1) integrating GLoRA into the transformer blocks for efficient fine-tuning, and (2) optimizing GLoRA with two complementary losses. The original diffusion loss leverages reference images to preserve appearance details, and the AiT Loss utilizes reference videos to focus on motion patterns. The pre-trained text-to-video model remains frozen throughout training; only GLoRA parameters are updated. During inference, trained GLoRA weights are loaded, and customized videos are generated conditioned solely on the input prompt.}
    \label{fig:overall_framework}
    \vspace{-1em}
\end{figure*}

\subsection{Text-to-Video Generation}

Text-to-video generation has progressed from early GAN-based~\cite{conditionalgan, mocogan} and autoregressive models~\cite{lvg, ccvs} to diffusion-based methods~\cite{makeavideo, videoldm, text2videozero}, which now dominate in quality and controllability. Two diffusion architectures prevail. UNet-based models such as AnimateDiff~\cite{animatediff} introduce motion modules into text-to-image pipelines; ModelScopeT2V~\cite{modelscope}, ZeroScope~\cite{zeroscope}, and Stable Video Diffusion~\cite{stablevideodiffusion} enhance fidelity and realism via spatiotemporal attention and multi-stage training. Diffusion Transformer-based models like CogVideoX~\cite{2022cogvideo, 2024cogvideox}, Wan~\cite{wan}, and HunyuanVideo~\cite{hunyuanvideo} leverage hierarchical training, adaptive fusion, and scalable pretraining with 3D VAE designs to achieve high-resolution, long-range, and semantically aligned generation. These developments provide a foundation for customized video synthesis.

\subsection{Customized Video Generation}

Customized video generation involves appearance control, motion control, or both. Appearance methods preserve identity from a few images~\cite{customvideo, customcrafter, dynvideo, clgc}, while motion methods transfer dynamics from reference videos~\cite{tuneavideo, motionshop, det}. Combined approaches often use LoRA or Adapters to separately tune spatial and temporal layers~\cite{motiondirector, customizeavideo}, sometimes enhanced with text inversion~\cite{dreamvideo}. However, stage-wise training suffers from concept interference. CustomTTT~\cite{customttt} uses test-time tuning to reduce domain shift but risks error accumulation. DualReal~\cite{dualreal} jointly trains appearance and motion adapters in CogVideoX~\cite{2022cogvideo} via a controller that fuses text and visual features during training. In inference, only text is available, causing the controller to assign higher weights to appearance, which often leads to weak motion customization. Moreover, to mitigate appearance leakage, some methods swap appearance components at inference~\cite{motiondirector, customizeavideo} or condition on an appearance image~\cite{dreamvideo}, yet robust disentanglement remains unresolved.

%% file: 3_method.tex
\section{Method}
\label{sec:method}

\begin{table*}[t]
    \centering
    \resizebox{\linewidth}{!}{
        \begin{tabular}{c|ccc|ccc|cc|cc|ccc}
            \hline
            \toprule
            \textbf{Models} & \multicolumn{3}{c|}{\underline{\textbf{\textit{Composite Metrics}}}} & \multicolumn{3}{c|}{\underline{\textbf{\textit{Semantic Alignment}}}} & \multicolumn{2}{c|}{\underline{\textbf{\textit{Motion Dynamism}}}} & \multicolumn{2}{c|}{\underline{\textbf{\textit{Temporal Consistency}}}} & \multicolumn{3}{c}{\underline{\textbf{\textit{Perceptual Quality}}}}  \\ 
            \textbf{(on CogVideoX~\cite{cogvideo})} & \textbf{ARS} $\downarrow$ & \textbf{NAS} $\uparrow$ & \textbf{AAS} $\uparrow$ & \textbf{CLIP-T} $\uparrow$ & \textbf{CLIP-I} $\uparrow$ & \textbf{Mot-Fid} $\uparrow$ & \textbf{Mot-Smth} $\uparrow$ & \textbf{Dyn-Deg} $\uparrow$ & \textbf{Subj-Con} $\uparrow$ & \textbf{Bkgd-Con} $\uparrow$ & \textbf{Pick-Sc} $\uparrow$ & \textbf{Aesth-Q} $\uparrow$ & \textbf{Img-Q} $\uparrow$ \\ \midrule
        \textbf{CogVideoX-5B~\cite{cogvideo}} & 3.10  & 91.82  & 65.70  & \colorbox{secondpurple}{33.03}  & 61.18  & 34.76  & 98.52  & \colorbox{bestred}{93.81}  & 93.22  & 95.60  & \colorbox{secondpurple}{20.94}  & \colorbox{secondpurple}{62.32}  & 63.59   \\ 
        \textbf{+ LoRA~\cite{cogvideo}} & 3.70  & 86.76  & 61.46  & 31.39  & 65.93  & 39.74  & \colorbox{secondpurple}{98.78}  & 48.57  & 94.70  & \colorbox{bestred}{95.97}  & 20.25  & 58.32  & 60.90   \\ \midrule
        \textbf{DualReal-Cog5B~\cite{dualreal}} & \colorbox{bestred}{1.90}  & 92.04  & 64.95  & \colorbox{bestred}{33.36}  & \colorbox{bestred}{71.30}  & 42.75  & 98.30  & 56.67  & \colorbox{bestred}{95.70}  & \colorbox{secondpurple}{95.74}  & \colorbox{bestred}{20.97}  & \colorbox{bestred}{63.63}  & \colorbox{bestred}{71.10}   \\ \midrule
        \textbf{JointTuner-Cog2B} & \colorbox{secondpurple}{2.90}  & \colorbox{secondpurple}{93.28}  & \colorbox{secondpurple}{66.35}  & 31.32  & 67.71  & \colorbox{bestred}{70.16}  & \colorbox{bestred}{98.98}  & 64.05  & \colorbox{secondpurple}{95.41}  & 95.17  & 20.53  & 59.58  & 60.65   \\ 
        \textbf{JointTuner-Cog5B} & 3.40  & \colorbox{bestred}{94.66}  & \colorbox{bestred}{67.75}  & 31.16  & \colorbox{secondpurple}{68.55}  & \colorbox{secondpurple}{62.15}  & 97.73  & \colorbox{secondpurple}{80.22}  & 94.80  & 95.09  & 20.35  & 59.33  & \colorbox{secondpurple}{68.11}   \\
        \bottomrule
        \hline
    \end{tabular}
    }
    \vspace{-1em}
    \caption{Quantitative comparison of DiT-based methods, built on CogVideoX. We highlight the \colorbox{bestred}{best} and \colorbox{secondpurple}{second-best} values for each metric.}
    \label{tab:overall_performance_cog}
    \vspace{-1em}
\end{table*}

\begin{table}[t]
    \centering
     \resizebox{\linewidth}{!}{
        \begin{tabular}{c|c|cccc}
        \hline
        \toprule
        \textbf{Models} & \multicolumn{5}{c}{\textbf{Example: \textit{A clock driving in the snow.}}} \\ \cline{2-6}
        \textbf{(on CogVideoX~\cite{cogvideo})} & \textbf{AAS} $\uparrow$& \textbf{CLIP-T} $\uparrow$& \textbf{CLIP-I} $\uparrow$& \textbf{Mot-Fid} $\uparrow$ & \textbf{Img-Q} $\uparrow$ \\ \midrule
        \textbf{CogVideoX-5B~\cite{cogvideo}} & 68.52  & \colorbox{bestred}{33.87}  & 58.87  & 34.11  & \colorbox{bestred}{75.61}  \\ 
        \textbf{+LoRA~\cite{cogvideo}} & 57.96  & 31.13  & \colorbox{bestred}{68.75}  & 41.63  & 59.28  \\ \hline
        \textbf{DualReal-Cog5B~\cite{dualreal}} & 70.64  & 31.15  & \colorbox{secondpurple}{63.90}  & 54.70  & 72.83  \\ \hline
        \textbf{JointTuner-Cog2B} & \colorbox{secondpurple}{71.53}  & \colorbox{secondpurple}{31.81}  & 56.61  & \colorbox{secondpurple}{67.43}  & \colorbox{secondpurple}{74.65}  \\ 
        \textbf{JointTuner-Cog5B} & \colorbox{bestred}{72.25}  & 30.17  & 56.28  & \colorbox{bestred}{91.35}  & 73.88  \\  \bottomrule \hline
    \end{tabular}
    }
    \vspace{-1em}
    \caption{Partial quantitative results corresponding to the customized video in Fig.~\ref{fig:overall_visualization_cog}. We highlight the \colorbox{bestred}{best} and \colorbox{secondpurple}{second-best} values for each metric.}
    \label{tab:evaluate_sample_cog}
    \vspace{-1em}
\end{table}

In this section, we introduce the concepts of text-to-video diffusion models and low-rank adaptation, followed by the proposed JointTuner framework.

\subsection{Preliminaries}
\label{sec:preliminaries}

\paragraph{Text-to-Video Diffusion Models} Following previous customized video generation methods~\cite{motiondirector, dreamvideo}, we adopt a diffusion-based framework where the denoising network \(\epsilon_\theta\) is trained to predict additive Gaussian noise from a noisy video latent \(z_t\), conditioned on a text prompt \(y\). We explore two architectural backbones:
(1) \textit{UNet Backbone.} A 3D UNet~\cite{diffusion} adopts a U-shaped encoder and decoder structure that combines convolutional blocks and Transformer blocks. It integrates spatial Transformers for per-frame semantics and temporal Transformers for inter-frame dynamics. 
(2) \textit{DiT Backbone.} Diffusion Transformer (DiT)~\cite{dit} replaces all the convolutional layers with a fully attention-based design. It allows long-range dependency modeling, improves capacity on high-resolution content, and naturally extends to video by applying spatiotemporal attention. The network ($\epsilon_\theta$) is optimized with a text encoder ($\tau_\theta$) via:
\begin{equation}
    \mathcal{L}_{z_0,y,\epsilon,t}\left[ \epsilon, \epsilon_\theta(z_t, t, \tau_\theta(y)) \right],
    \label{eq:generic_denoise_loss}
\end{equation}
where $z_0 \in \mathbb{R}^{F \times C \times H \times W }$ denotes the latent code of the video, and $F, C, H, W$ denote the number of frames, channels, height, and width of the latent code, respectively. $y$ is the text prompt, $\epsilon\sim\mathcal{N}(0,I)$ represents Gaussian noise, and $t\sim\mathcal{U}(0,T)$ corresponds to the time steps. A typical $\mathcal{L}(\cdot)$ is Mean Squared Error (MSE) Loss, formally as follows:
\begin{equation}
  \mathbb{E}_{z_0,y,\epsilon,t}\left[\|\epsilon-\epsilon_\theta(z_t,t,\tau_\theta(y))\|_2^2\right].
  \label{eq:mse_diffusion_loss}
\end{equation}

\paragraph{Low-Rank Adaptation (LoRA)} LoRA~\cite{lora} enables parameter-efficient adaptation by decomposing weight updates as $W = W_0 + BA$, where $W_0 \in \mathbb{R}^{d \times k}$ is frozen, and $B \in \mathbb{R}^{d \times r}$ and $A \in \mathbb{R}^{r \times k}$ (with $r \ll d, k$) are trainable low-rank matrices. This design significantly reduces the number of trainable parameters while preserving model capacity. The forward pass is given by
\begin{equation}
x^\prime = W_0x + \alpha \cdot BAx,
\label{eq:generic_lora_propagation}
\end{equation}
where $x$ is the input feature and $x^\prime$ is the adapted output feature, and $\alpha$ is a fixed scaling factor.

\subsection{Gated Low-Rank Adaptation}

Concept interference arises in stage-wise paradigms because fine-tuned parameters learned in separate stages are not effectively fused during inference. To resolve this, a unified joint training framework is needed. However, applying conventional LoRA to joint fine-tuning leads to suboptimal performance due to conflicts between shared LoRA modules adapting to distinct modalities~\cite{dreamvideo, motiondirector}.

To mitigate such conflicts, each LoRA can be trained as a modality-specific expert module, dedicated either to appearance or motion. We propose \textbf{Gated Low-Rank Adaptation (GLoRA)}, a novel architecture that introduces a context-aware linear activation layer, termed the gating regulator, to dynamically modulate the activation state of each LoRA based on input modality. The forward propagation formulation of GLoRA is defined as follows:
\begin{equation}
x^\prime = W_0x + \underbrace{\mathrm{Mean}(\sigma(Gx))}_{\text{gating regulation}} \cdot BAx,
\end{equation}
where $G$ is a learnable projection, $\sigma$ is the sigmoid function, and $\mathrm{Mean}(\cdot)$ ensures stability across frames and batches.

The gating regulator receives the input to the associated layer, which carries rich spatial and temporal semantics, serving as a natural context signal. During training, by alternating between appearance-focused and motion-focused reconstruction tasks, the gating regulators learn to selectively activate or inhibit their corresponding LoRA according to whether they specialize in appearance or motion. At inference time, this gating mechanism enables dynamic composition of semantically specialized LoRAs, effectively suppressing concept interference and enabling coherent fusion.

\begin{figure}[t]
  \centering
    \includegraphics[width=\linewidth]{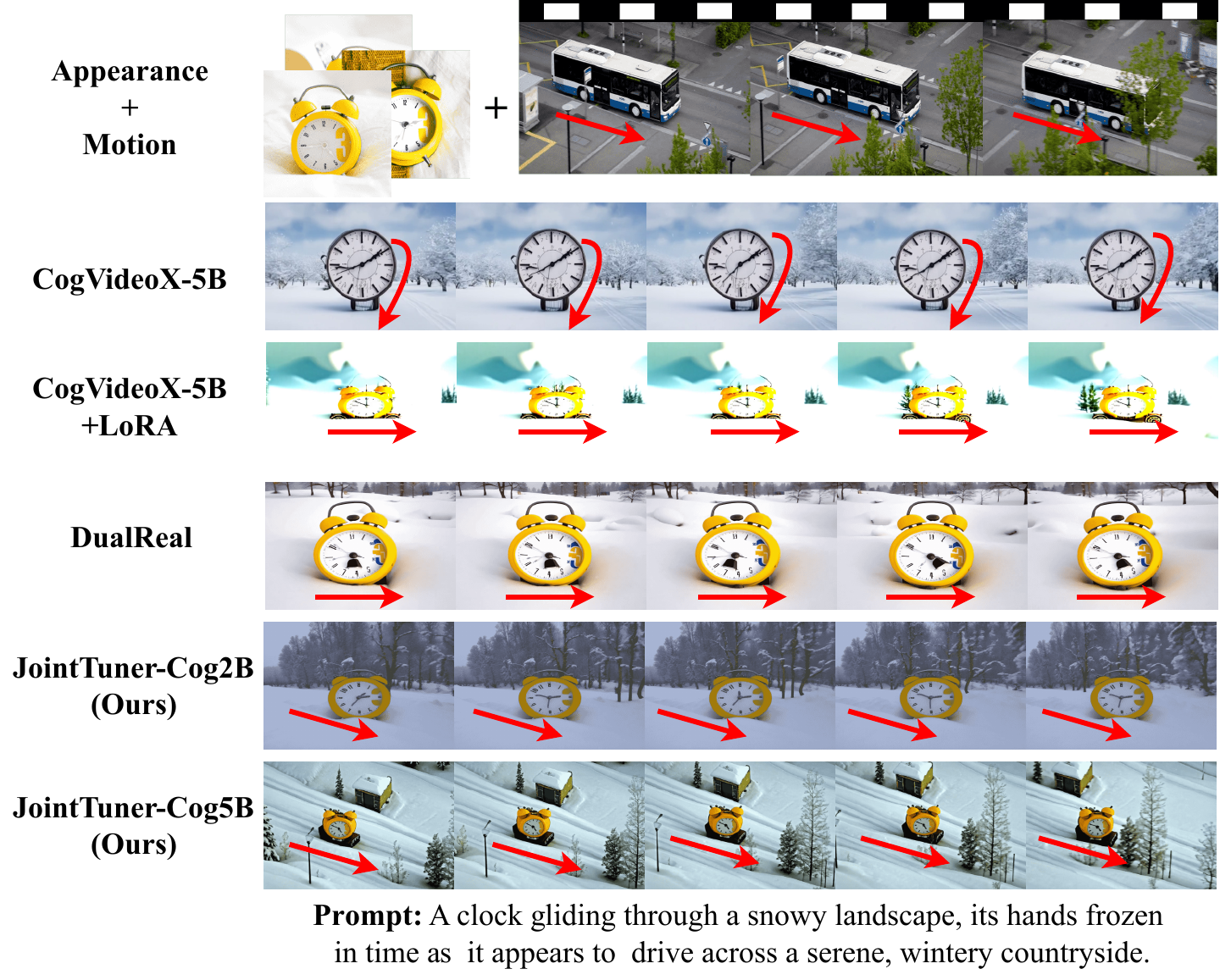}
    \vspace{-2em}
    \captionof{figure}{Qualitative results of appearance-motion combined customization on CogVideoX.}
    \label{fig:overall_visualization_cog}
    \vspace{-1em}
\end{figure}

\begin{table*}[t]
    \centering
    \resizebox{\linewidth}{!}{
        \begin{tabular}{c|ccc|ccc|cc|cc|ccc}
            \hline
            \toprule
            \textbf{Models} & \multicolumn{3}{c|}{\underline{\textbf{\textit{Composite Metrics}}}} & \multicolumn{3}{c|}{\underline{\textbf{\textit{Semantic Alignment}}}} & \multicolumn{2}{c|}{\underline{\textbf{\textit{Motion Dynamism}}}} & \multicolumn{2}{c|}{\underline{\textbf{\textit{Temporal Consistency}}}} & \multicolumn{3}{c}{\underline{\textbf{\textit{Perceptual Quality}}}}  \\ 
            \textbf{(on ZeroScope~\cite{zeroscope})} & \textbf{ARS} $\downarrow$ & \textbf{NAS} $\uparrow$ & \textbf{AAS} $\uparrow$ & \textbf{CLIP-T} $\uparrow$ & \textbf{CLIP-I} $\uparrow$ & \textbf{Mot-Fid} $\uparrow$ & \textbf{Mot-Smth} $\uparrow$ & \textbf{Dyn-Deg} $\uparrow$ & \textbf{Subj-Con} $\uparrow$ & \textbf{Bkgd-Con} $\uparrow$ & \textbf{Pick-Sc} $\uparrow$ & \textbf{Aesth-Q} $\uparrow$ & \textbf{Img-Q} $\uparrow$ \\ \midrule
        \textbf{MotionDirector-ZS~\cite{motiondirector}} & \colorbox{secondpurple}{2.00}  & \colorbox{secondpurple}{92.78}  & \colorbox{secondpurple}{62.71}  & \colorbox{secondpurple}{28.56}  & \colorbox{bestred}{70.73}  & 70.12  & 96.66  & 34.22  & \colorbox{bestred}{94.12}  & \colorbox{bestred}{94.49}  & \colorbox{secondpurple}{20.37}  & \colorbox{secondpurple}{53.27}  & \colorbox{secondpurple}{64.62}  \\ 
        \textbf{DreamVideo-ZS~\cite{dreamvideo}} & 2.50  & 92.27  & 62.22  & 27.39  & 51.70  & \colorbox{secondpurple}{72.60}  & \colorbox{secondpurple}{96.94}  & \colorbox{bestred}{61.56}  & 89.92  & \colorbox{secondpurple}{94.10}  & 20.01  & 48.66  & 59.33  \\  \midrule
        \textbf{JointTuner-ZS} & \colorbox{bestred}{1.50}  & \colorbox{bestred}{96.14}  & \colorbox{bestred}{64.15}  & \colorbox{bestred}{32.14}  & \colorbox{secondpurple}{62.44}  & \colorbox{bestred}{72.61}  & \colorbox{bestred}{97.78}  & \colorbox{secondpurple}{46.89}  & \colorbox{secondpurple}{91.92}  & 93.83  & \colorbox{bestred}{21.19}  & \colorbox{bestred}{57.14}  & \colorbox{bestred}{65.53}  \\ 
        \bottomrule
        \hline
    \end{tabular}
    }
    \vspace{-1em}
    \caption{Quantitative comparison of UNet-based methods, built on ZeroScope. We highlight the \colorbox{bestred}{best} and \colorbox{secondpurple}{second-best} values for each metric.}
    \label{tab:overall_performance_zs}
    \vspace{-1em}
\end{table*}

\subsection{Appearance-independent Temporal Loss}

In customized video generation, motion reference videos often intertwine spatial appearance with temporal motion, leading to appearance leakage during fine-tuning. To address this, we propose a shift perturbation strategy that introduces structured noise along specific feature dimensions to disentangle motion and appearance representations. As illustrated in Fig.~\ref{fig:different_shift_visual}, we apply three types of perturbations: channel-wise $(F, C{=}1, H, W)$, spatial-wise $(F, C, H{=}1, W{=}1)$, and temporal-wise $(F{=}1, C, H, W)$. Channel-wise and temporal-wise shifts effectively disrupt spatial correlations while preserving coherent motion trajectories, whereas spatial-wise shifts mainly modify color tones and degrade motion consistency.

Given prior findings that motion is encoded in high-frequency components, while appearance resides in low-frequency components~\cite{freelong, riflex}, we further conduct a spectral analysis to explain the phenomena. Specifically, we perform a 3D Fast Fourier Transform on the denoised latents and normalize the energy distribution across frequency bands. The quantitative results, presented in Fig.~\ref{fig:different_shift_result}, show that channel-wise and temporal-wise shifts suppress low-frequency components and amplify high-frequency signals, confirming their effectiveness in isolating motion features. In contrast, spatial-wise shift loses effectiveness at higher timesteps, indicating limited ability to disentangle motion.

Building upon these insights, we propose the \textbf{Appearance-independent Temporal Loss (AiT Loss)} to enforce motion-focused learning. During fine-tuning, we inject shift noise $\Delta_{noise} \in \mathbb{R}^{1 \times 1 \times  H \times  W}$ that remains constant across both temporal and channel dimensions into the predicted noise. This structured perturbation disrupts appearance-related cues such as texture, color, and identity, compelling the diffusion model to rely solely on motion dynamics for reconstruction. Consequently, AiT Loss enhances motion fidelity and mitigates appearance contamination during customized video generation. The AiT Loss is formally defined as follows:
\begin{equation}
\mathcal{L}{z_0,y,\epsilon,t}\left[ \epsilon, (\epsilon_{\theta, \pi}(z_t, t, \tau_\theta(y)) + \Delta_{noise}) \right],
\label{eq:ait_loss}
\end{equation}
where $\pi$ denotes the injected GLoRA parameters.

\subsection{Training and Inference}

The model is jointly trained with GLoRA by alternating between image and video inputs. Images are treated as single-frame videos, while videos are uniformly sampled into $N$ frames. Image inputs use the original denoising loss (Eq.~\ref{eq:generic_denoise_loss}) to preserve appearance, whereas video inputs adopt the AiT Loss (Eq.~\ref{eq:ait_loss}) to focus on motion patterns and suppress appearance contamination.

During inference, the trained GLoRA parameters are integrated into the pre-trained text-to-video diffusion model, enabling the generation of customized videos conditioned solely on the textual prompt. 

We provide the pseudocode for JointTuner  and the stage-wise strategy approach in Appendix~\ref{sec:algorithm_compare}, and compare the differences between their training and inference procedures.

%% file: 4_exp.tex
\section{Experiments}
\label{sec:exp}

\begin{figure}[t]
  \centering
    \includegraphics[width=\linewidth]{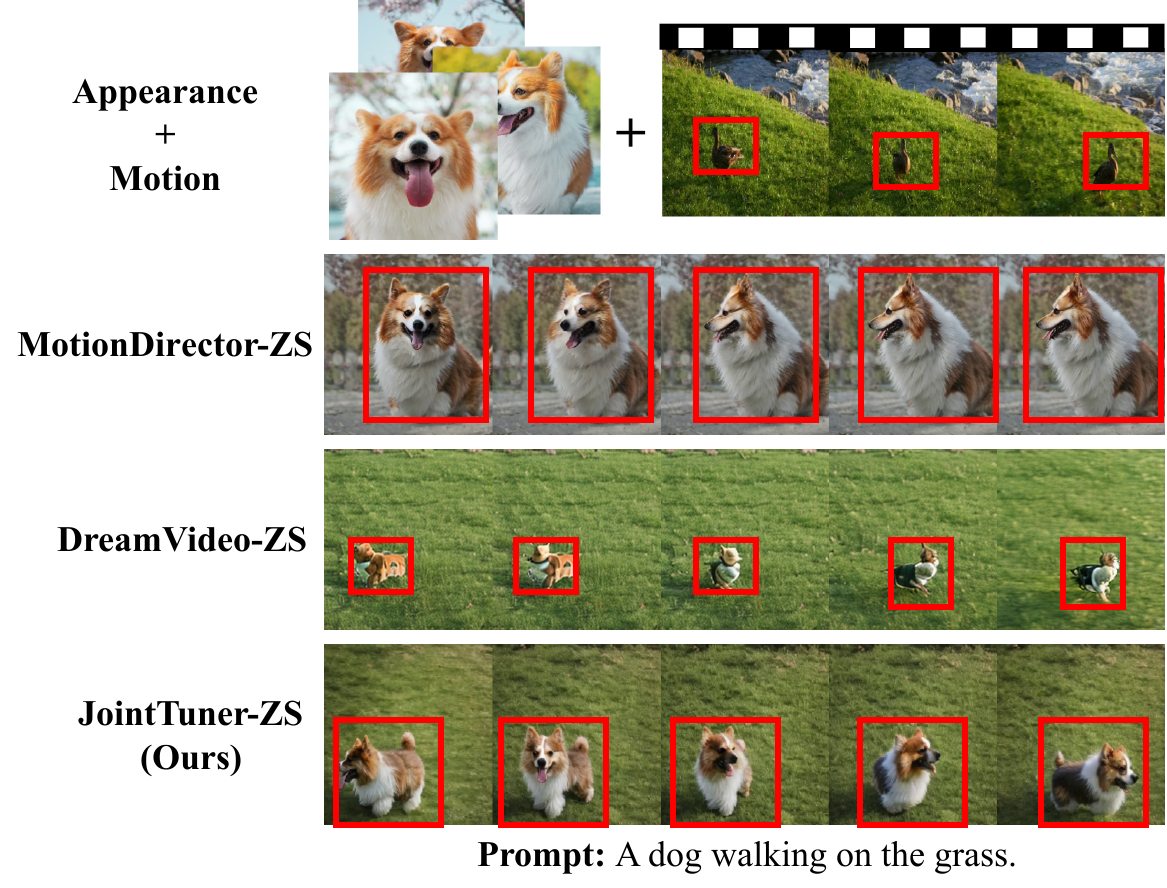}
    \vspace{-2em}
    \captionof{figure}{Qualitative results of appearance-motion combined customization on ZeroScope.}
    \label{fig:overall_visualization_zs}
    \vspace{-1em}
\end{figure}

This section presents the proposed benchmark, comparisons with state-of-the-art methods, ablation studies, and failure cases. More results, including hyperparameter analysis and further visualization results, are provided in Appendix~\ref{sec:more_results}.

\begin{table*}[t]
    \centering
    \resizebox{\linewidth}{!}{
        \begin{tabular}{c|ccc|ccc|cc|cc|ccc}
            \hline
            \toprule
            \textbf{Models} & \multicolumn{3}{c|}{\underline{\textbf{\textit{Composite Metrics}}}} & \multicolumn{3}{c|}{\underline{\textbf{\textit{Semantic Alignment}}}} & \multicolumn{2}{c|}{\underline{\textbf{\textit{Motion Dynamism}}}} & \multicolumn{2}{c|}{\underline{\textbf{\textit{Temporal Consistency}}}} & \multicolumn{3}{c}{\underline{\textbf{\textit{Perceptual Quality}}}}  \\ 
            \textbf{(on CogVideoX~\cite{cogvideo})} & \textbf{ARS} $\downarrow$ & \textbf{NAS} $\uparrow$ & \textbf{AAS} $\uparrow$ & \textbf{CLIP-T} $\uparrow$ & \textbf{CLIP-I} $\uparrow$ & \textbf{Mot-Fid} $\uparrow$ & \textbf{Mot-Smth} $\uparrow$ & \textbf{Dyn-Deg} $\uparrow$ & \textbf{Subj-Con} $\uparrow$ & \textbf{Bkgd-Con} $\uparrow$ & \textbf{Pick-Sc} $\uparrow$ & \textbf{Aesth-Q} $\uparrow$ & \textbf{Img-Q} $\uparrow$ \\ \midrule
        \textbf{CogVideoX-5B} & \colorbox{bestred}{2.30}  & \colorbox{secondpurple}{93.49}  & \colorbox{secondpurple}{65.70}  & \colorbox{bestred}{33.03}  & 61.18  & 34.76  & \colorbox{secondpurple}{98.52}  & \colorbox{bestred}{93.81}  & 93.22  & \colorbox{secondpurple}{95.60}  & \colorbox{bestred}{20.94}  & \colorbox{bestred}{62.32}  & 63.59   \\ 
        \textbf{Cog5B + LoRA} & \colorbox{secondpurple}{2.60}  & 88.52  & 61.46  & \colorbox{secondpurple}{31.39}  & 65.93  & 39.74  & \colorbox{bestred}{98.78}  & 48.57  & 94.70  & \colorbox{bestred}{95.97}  & 20.25  & 58.32  & 60.90   \\ 
        \textbf{Cog5B + GLoRA} & 2.80  & 91.05  & 62.90  & 30.48  & \colorbox{bestred}{69.28}  & \colorbox{secondpurple}{55.06}  & 97.80  & 40.00  & \colorbox{bestred}{95.13}  & 95.43  & 20.13  & 57.76  & \colorbox{secondpurple}{67.91}   \\ 
        \textbf{Cog5B + GLoRA + AiT  (Ours)} & \colorbox{bestred}{2.30}  & \colorbox{bestred}{96.89}  & \colorbox{bestred}{67.75}  & 31.16  & \colorbox{secondpurple}{68.55}  & \colorbox{bestred}{62.15}  & 97.73  & \colorbox{secondpurple}{80.22}  & \colorbox{secondpurple}{94.80}  & 95.09  & \colorbox{secondpurple}{20.35}  & \colorbox{secondpurple}{59.33}  & \colorbox{bestred}{68.11}   \\ \bottomrule 
        \hline
    \end{tabular}
    }
    \vspace{-1em}
    \caption{Quantitative ablation results for each core component of the proposed JointTuner, built on CogVideoX-5B. We highlight the \colorbox{bestred}{best} and \colorbox{secondpurple}{second-best} values for each metric.}
    \label{tab:ablation_cog5b}
    \vspace{-1em}
\end{table*}

\subsection{Benchmark}

Most existing appearance-motion customization methods~\cite{motiondirector, customttt, dualreal} rely on researcher-curated datasets and emphasize appearance-oriented metrics (e.g., CLIP-Image, Subject Consistency), while often overlooking motion-specific evaluation. To address this, we construct a comprehensive benchmark. Below is a brief introduction of the benchmark, and more details can be found in Appendix~\ref{sec:benchmark}.

\textbf{Dataset.} We construct a comprehensive dataset of 90 appearance-motion combinations using public sources, comprising 15 subjects from CustomDiffusion~\cite{customdiffusion}, DreamBooth~\cite{dreambooth}, and MotionDirector~\cite{motiondirector}, along with nine motion types from DAVIS2016~\cite{davis2016} and three composite motions (each composed of five video clips) from UCF101~\cite{ucf101}. The dataset includes six distinct subject-motion pairing strategies and is evaluated under five specific background settings: grassland, jungle, snowscape, beach, and cobblestone street. This yields a benchmark of 450 customized videos covering diverse combinations of subjects, motions, and backgrounds.

\textbf{Metrics.} We evaluate generation quality across four dimensions:
(1) \textit{Semantic alignment} includes CLIP-Text (CLIP-T) for text-video alignment, CLIP-Image (CLIP-I) for image-video correspondence, and Motion-Fidelity (Mot-Fid), which measures motion pattern consistency using trajectory tracking from CoTracker3~\cite{cotracker3}.
(2) \textit{Motion dynamism} is quantified by two metrics:
Motion-Smoothness (Mot-Smth), which assesses temporal coherence via AMT-based interpolation priors, and
Dynamic-Degree (Dyn-Deg), which measures motion intensity using optical flow magnitude from RAFT~\cite{raft}.
Both are implemented in VBench~\cite{vbench}.
(3) \textit{Temporal consistency} includes Subject-Consistency (Subj-Con) and Background-Consistency (Bkgd-Con), evaluated using DINO and CLIP feature similarity across frames.
(4) \textit{Perceptual quality} aggregates Pick-Score (Pick-Sc), Aesthetic-Quality (Aesth-Q), and Imaging-Quality (Img-Q) at the frame level.
To holistically compare methods, we report three \textbf{composite metrics}: the Average Rank Score (ARS), which represents the mean rank of a model across all individual metrics where a lower rank indicates better performance; the Normalized Average Score (NAS), in which each metric is normalized so that the best-performing method receives a score of 100 and NAS is the average of these normalized values; and the Absolute Average Score (AAS), defined as the arithmetic mean of raw metric values.

\subsection{Main Results}

\begin{figure}[t]
  \centering
  \includegraphics[width=\linewidth]{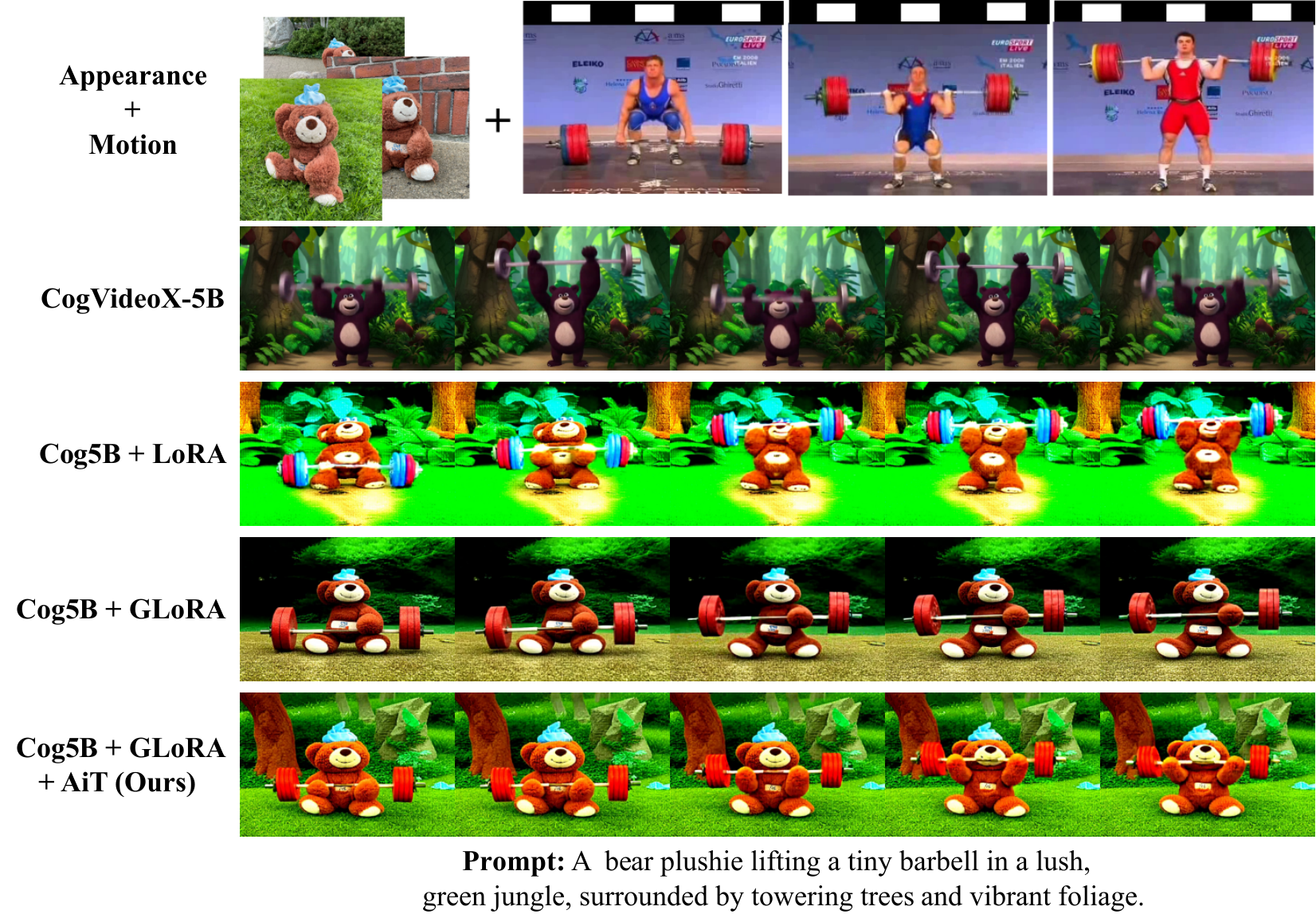}
  \vspace{-2em}
  \caption{Qualitative ablation results for each component of JointTuner based on CogVideoX-5B.}
  \label{fig:ablation_cog5b}
  \vspace{-1em}
\end{figure}

We compare JointTuner with state-of-the-art methods. For clarity, ``-ZS'', ``-Cog2B'', and ``-Cog5B'' indicate variants built on ZeroScope~\cite{zeroscope}, CogVideoX-2B~\cite{cogvideo}, and CogVideoX-5B~\cite{cogvideo}, respectively.  Implementation details of JointTuner and experimental settings for the comparison methods are provided in Appendix~\ref{sec:exp_setting}.

\textbf{Quantitative results.} As shown in Tab.~\ref{tab:overall_performance_cog}, CogVideoX-5B exhibits poor semantic alignment, and naive joint tuning via typical LoRA yields only marginal improvements. This indicates that simply injecting LoRA for joint fine-tuning also suffers from concept interference, leading to poor performance in combined customization. DualReal achieves good performance in appearance-related scores (high CLIP-I), but the controller that fuses appearance and motion suffers from a gap between training and inference, resulting in poor motion scores (low Mot-Fid and Dyn-Deg). In contrast, JointTuner-Cog2B achieves good performance on both CLIP-I and Mot-Fid scores, striking a better balance. JointTuner-Cog5B further improves CLIP-I and Img-Q, benefiting from the stronger backbone. On ZeroScope (Tab.~\ref{tab:overall_performance_zs}), JointTuner leads in perceptual quality and consistently ranks within the top-2 across most metrics. While MotionDirector excels in static fidelity, it underperforms in motion capture; conversely, DreamVideo achieves high motion scores but suffers from degraded appearance. Additional experiments on ModelScope~\cite{modelscope} corroborate these trends and are reported in Appendix~\ref{sec:more_results}. This indicates that stage-wise paradigms, due to concept interference, fail to achieve strong performance in both appearance and motion.

\textbf{Qualitative results.} As illustrated in Fig.~\ref{fig:overall_visualization_cog}, CogVideoX-5B, without any fine-tuning, exhibits no customization in either appearance or motion. Simply injecting LoRA for joint fine-tuning suffers from concept interference, achieving only appearance customization. DualReal fails to generate the desired motion trajectories due to the gap between training and inference. In contrast, JointTuner, through a carefully designed joint training strategy, successfully achieves effective integration of both appearance and motion. Similarly, in Fig.~\ref{fig:overall_visualization_zs}, JointTuner produces semantically aligned outputs, whereas MotionDirector yields visually sharp but static frames, and DreamVideo captures dynamic motion at the cost of identity consistency.

\textbf{Metrics analysis.} To assess the effectiveness of our comprehensive evaluation metric, we conducted a case study using the example in Fig.~\ref{fig:overall_visualization_cog}, with quantitative results in Tab.~\ref{tab:evaluate_sample_cog}. Based on prior appearance-related metrics, both direct LoRA injection with joint fine-tuning and DualReal achieve high scores. However, as shown in Fig.~\ref{fig:overall_visualization_cog}, neither method successfully generates the desired motion trajectories. Therefore, it is necessary to introduce motion-related metrics to quantitatively demonstrate this limitation. Furthermore, visual results show that our JointTuner effectively integrates appearance and motion customization and achieves the best overall evaluation scores, reflecting the consistency between our proposed comprehensive evaluation metrics and the qualitative results.

\subsection{Ablation Studies}

\begin{table*}[t]
    \centering
    \resizebox{\linewidth}{!}{
        \begin{tabular}{c|ccc|ccc|cc|cc|ccc}
            \hline
            \toprule
            \textbf{Models} & \multicolumn{3}{c|}{\underline{\textbf{\textit{Composite Metrics}}}} & \multicolumn{3}{c|}{\underline{\textbf{\textit{Semantic Alignment}}}} & \multicolumn{2}{c|}{\underline{\textbf{\textit{Motion Dynamism}}}} & \multicolumn{2}{c|}{\underline{\textbf{\textit{Temporal Consistency}}}} & \multicolumn{3}{c}{\underline{\textbf{\textit{Perceptual Quality}}}}  \\ 
            \textbf{(on ZeroScope~\cite{zeroscope})} & \textbf{ARS} $\downarrow$ & \textbf{NAS} $\uparrow$ & \textbf{AAS} $\uparrow$ & \textbf{CLIP-T} $\uparrow$ & \textbf{CLIP-I} $\uparrow$ & \textbf{Mot-Fid} $\uparrow$ & \textbf{Mot-Smth} $\uparrow$ & \textbf{Dyn-Deg} $\uparrow$ & \textbf{Subj-Con} $\uparrow$ & \textbf{Bkgd-Con} $\uparrow$ & \textbf{Pick-Sc} $\uparrow$ & \textbf{Aesth-Q} $\uparrow$ & \textbf{Img-Q} $\uparrow$ \\ \midrule
        \textbf{Spatial Shift} & 2.40  & 93.28  & 60.69  & 31.13  & 59.80  & 59.50  & \colorbox{bestred}{98.06}  & 34.89  & \colorbox{bestred}{92.23}  & \colorbox{bestred}{94.17}  & 20.80  & 52.82  & 63.54   \\ 
        \textbf{Full Shift} & \colorbox{secondpurple}{2.10}  & \colorbox{secondpurple}{99.11}  & \colorbox{secondpurple}{64.06}  & \colorbox{secondpurple}{31.19}  & \colorbox{secondpurple}{62.07}  & \colorbox{bestred}{75.11}  & 97.54  & \colorbox{bestred}{47.56}  & 91.70  & 93.54  & \colorbox{secondpurple}{21.05}  & \colorbox{secondpurple}{56.45}  & \colorbox{secondpurple}{64.43}   \\ \midrule
        \textbf{Channel-Temp Shift (Ours)} & \colorbox{bestred}{1.50 } & \colorbox{bestred}{99.43}  & \colorbox{bestred}{64.15}  & \colorbox{bestred}{32.14}  & \colorbox{bestred}{62.44}  & \colorbox{secondpurple}{72.61}  & \colorbox{secondpurple}{97.78}  & \colorbox{secondpurple}{46.89}  & \colorbox{secondpurple}{91.92}  & \colorbox{secondpurple}{93.83}  & \colorbox{bestred}{21.19}  & \colorbox{bestred}{57.14}  & \colorbox{bestred}{65.53}   \\ \bottomrule \hline
    \end{tabular}
    }
    \vspace{-1em}
    \caption{Quantitative results of different noise shift designs on ZeroScope. We highlight the \colorbox{bestred}{best} and \colorbox{secondpurple}{second-best} values for each metric.}
    \label{tab:shift_shape}
    \vspace{-1em}
\end{table*}

We perform ablation studies on two video generation backbones, CogVideoX-5B (DiT-based) and ZeroScope (UNet-based). Additional results, including the effect of the gating regulator in GLoRA and qualitative and quantitative analyses based on ZeroScope, are provided in Appendix~\ref{sec:more_results}.

\begin{figure}[t]
  \centering
  \includegraphics[width=\linewidth]{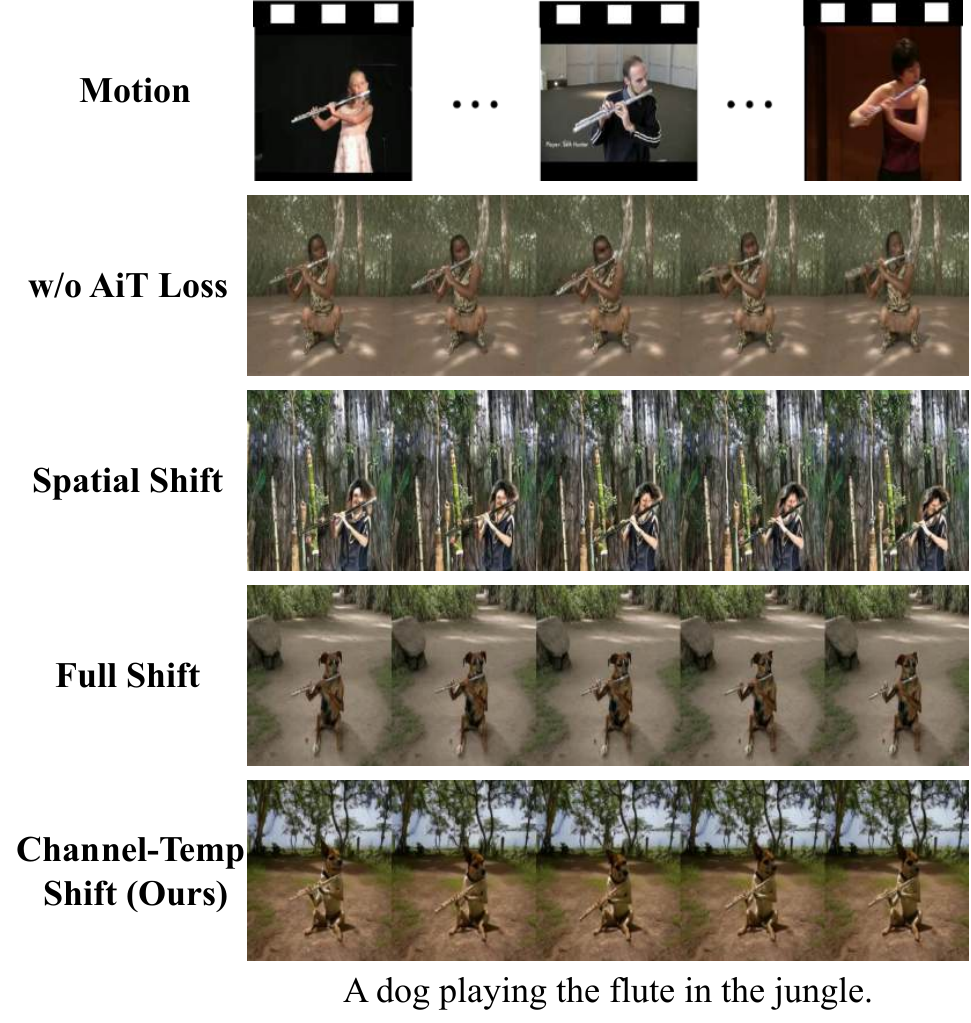}
  \vspace{-2em}
  \caption{Motion training outcomes using three noise shift strategies in the AiT Loss based on ZeroScope.}
  \label{noise_shape_sample}
  \vspace{-1em}
\end{figure}

\textbf{Quantitative ablations on CogVideoX-5B.}
We conduct quantitative ablation experiments on CogVideoX-5B to assess the impact of GLoRA and AiT Loss within the DiT backbone. As shown in Tab.~\ref{tab:ablation_cog5b}, integrating GLoRA leads to consistent improvements over standard LoRA across all key metrics, including NAS, AAS, CLIP-I, and Mot-Fid. The gains are particularly pronounced in appearance consistency and motion realism, demonstrating the effectiveness of the gating regulator. Furthermore, when AiT Loss is incorporated during training, we observe additional performance gains, especially in text-to-video alignment and video quality, reflected in higher CLIP-I and improved perceptual quality metrics. These results confirm that AiT Loss effectively mitigates appearance contamination while preserving target visual semantics.

\textbf{Qualitative ablations on CogVideoX-5B.} Illustrated in Fig.~\ref{fig:ablation_cog5b}, models equipped with GLoRA generate videos with significantly enhanced appearance and motion fidelity. During training, the model learns to dynamically activate appearance- or motion-specialized LoRA modules based on input semantics; at inference time, this enables coherent fusion of both features, preserving subject identity while producing natural motion. Moreover, the inclusion of AiT Loss further refines output quality, particularly in background rendering and adherence to textual prompts, without compromising motion dynamics.

\textbf{Impact of noise shift designs.} We compare three noise shift strategies for AiT Loss in ZeroScope-based implementation: Spatial Shift $(F,C,H{=}1,W{=}1)$, Full Shift $(F,C,H,W)$, and Channel-Temp Shift $(F{=}1,C{=}1,H,W)$. Tab.~\ref{tab:shift_shape} shows that Channel-Temp Shift achieves the best performance in semantic alignment and perceptual quality. Visuals in Fig.~\ref{noise_shape_sample} confirm it effectively suppresses appearance interference while preserving motion dynamics. In contrast, Spatial Shift suffers from severe appearance contamination, leading to distorted subject identities. Full Shift mitigates this issue to some extent, but still struggles with degraded background quality.

\begin{figure}[t]
  \centering
  \includegraphics[width=0.9\linewidth]{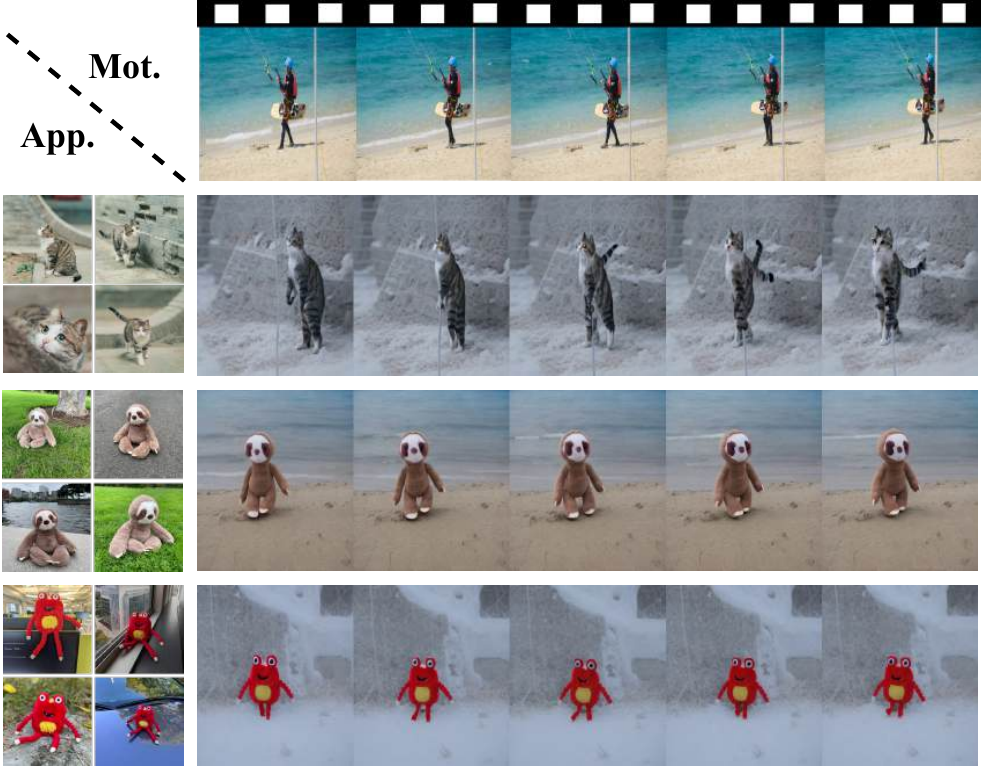}
  \vspace{-1em}
  \caption{Failure cases of our JointTuner based on ZeroScope, illustrating the results on different subjects: a cat, a sloth plushie, and a monster toy walking.}
  \label{fig:failure_sample}
  \vspace{-1em}
\end{figure}

\subsection{Failure Cases and Limitations}
\label{sec:failure_cases}

Despite the overall strong performance, our JointTuner may fail under extreme views. As illustrated in Fig.~\ref{fig:failure_sample}, the system occasionally encounters issues when the reference and target viewpoints differ significantly. For example, the side-facing cat aligns well with motion, while the sloth plushie and monster toy fail to maintain structural integrity due to the lack of side-view references. These cases highlight the limitations of appearance-motion composition when training data does not sufficiently cover certain viewpoints.

%% file: 5_conclusion.tex
\section{Conclusion}

In this work, we propose JointTuner, an adaptive joint training framework for appearance-motion combined customization. By integrating Gated LoRA and AiT Loss, JointTuner effectively mitigates concept interference that arises in stage-wise tuning and alleviates appearance contamination from reference videos. It is compatible with both UNet and DiT architectures, and demonstrates a strong ability to balance alignment between appearance and motion. To enable fair comparisons, we also introduce a comprehensive evaluation benchmark for combined customization. The benchmark includes 90 appearance-motion combinations and uses 10 automatic metrics in four dimensions, providing a foundation for assessing combined customization. Future work will explore strategies to augment reference priors and integrate 3D-aware representations to further improve robustness and generalization in appearance-motion combined customization.

%% file: X_suppl.tex
\renewcommand{\thetable}{A\arabic{table}}
\renewcommand{\thefigure}{A\arabic{figure}}

\clearpage
\setcounter{page}{1}
\setcounter{table}{0}
\setcounter{figure}{0}
\appendix
\maketitlesupplementary

This supplementary material is organized into the following sections:

\begin{itemize}
    \item \textbf{Algorithmic Comparisons}:  
    Presents pseudocode and detailed training and inference procedures for JointTuner and the stage-wise training paradigm, highlighting differences in parameter handling and optimization flow.

    \item \textbf{Benchmark Details}:  
    Describes the construction of our proposed benchmark, including the selection of subject and motion, data preprocessing strategies for different model architectures, and the design of fine-grained evaluation metrics across multiple quality dimensions.

    \item \textbf{Experiment Settings}:  
    Provides hyperparameters, training configurations, and hardware settings for JointTuner and compared methods.

    \item \textbf{Extended Analysis and Additional Results}:  
    Includes extended quantitative and qualitative results, ablation studies, analysis of the gating regulator in our GLoRA, sensitivity to key hyperparameters, and additional customized videos that demonstrate the appearance-motion combined customization capabilities of JointTuner in comparison with state-of-the-art approaches.
\end{itemize}

\section{Algorithmic Comparisons}
\label{sec:algorithm_compare}

We present the detailed procedures of JointTuner and the stage-wise training paradigm in Algorithms~\ref{alg:jointtuner} and~\ref{alg:two-stage_cvg}, respectively. The stage-wise approach consists of two distinct training phases: first, spatial LoRA parameters are optimized on image data to adapt the model's appearance representation; subsequently, temporal LoRA parameters are trained separately on video data to refine motion dynamics. During inference, both sets of parameters are applied concurrently. In contrast, JointTuner trains on mixed batches comprising both images and videos, employing modality-specific loss functions while updating a unified set of GLoRA parameters throughout the training process.

\begingroup
\begin{algorithm}
\SetKwData{InputData}{input}\SetKwData{OutputData}{output}
\SetKwFunction{Train}{Train}
\SetKwFunction{Generate}{Generate}
\SetKwInOut{Input}{Input}\SetKwInOut{Output}{Output}

\Input{Training data (images and videos), pretrained model weights}
\Output{Trained model, generated video}
\BlankLine
Initialize models: $vae$, $\tau_\theta$, $\epsilon_\theta$;
\BlankLine
Load pretrained model weights;
\BlankLine
Add GLoRA parameters $\pi$ to $\epsilon_\theta$ to obtain $\epsilon_{\theta, \pi}$;
\BlankLine
\textbf{Adaptive Joint Training}:
\BlankLine
 \For{each epoch}{
    \For{each batch in data loader}{
        Encode batch into latent space $z_0$ using $vae$; 
        \BlankLine
        Sample noise $\epsilon$ and add to $z_0$ to obtain $z_t$;
        \BlankLine
        Encode text embeddings $\tau_\theta(y)$;
        \BlankLine
        \uIf{batch is image}{
            Compute original diffusion loss with predicted noise:
            \[
            \mathcal{L}_{z_0,y,\epsilon,t}\left[ \epsilon, \epsilon_{\theta, \pi}(z_t, t, \tau_\theta(y)) \right];
            \]
        }
        \uElseIf{batch is video}{
            Compute AiT loss with predicted noise:
            \[
            \mathcal{L}_{z_0,y,\epsilon,t}\left[ \epsilon, (\epsilon_{\theta, \pi}(z_t, t, \tau_\theta(y))  + \Delta_{noise}) \right];
            \]
        }
        Update GLoRA parameters $\pi$;
    }
}
\BlankLine 
Save trained GLoRA checkpoints;
\BlankLine            
\textbf{Inference}:
\BlankLine 
Initialize pipeline with pretrained models;
\BlankLine 
Load trained GLoRA parameters $\pi$;
\BlankLine 
\For{each prompt}{
    Prepare input latents (random noise) $\hat{z}_t$;
    \BlankLine
    Perform the denoising process step by step, ultimately yielding the denoised latent $\hat{z}_0$:
    \[
    \hat{z}_{t-1} = \hat{z}_t - \hat{\epsilon}_{\theta, \pi}(\hat{z}_t, t, \tau_\theta(y));
    \]
    \BlankLine
    Decode denoised latent $\hat{z}_0$ into video using $vae$;
}
\caption{\textbf{JointTuner}}
\label{alg:jointtuner}
\end{algorithm}
\endgroup

\begingroup
\begin{algorithm}
\SetKwData{InputData}{input}\SetKwData{OutputData}{output}
\SetKwFunction{Initialize}{Initialize}
\SetKwFunction{Load}{Load}
\SetKwFunction{Freeze}{Freeze}
\SetKwFunction{AddTuningParameters}{AddTuningParameters}
\SetKwInOut{Input}{Input}\SetKwInOut{Output}{Output}

\Input{Training data (images and videos), pretrained model weights}
\Output{Trained model, generated video}
\BlankLine
Initialize models: $vae$, $\tau_\theta$, $\epsilon_\theta$;
\BlankLine 
Load pretrained model weights;
\BlankLine
\textbf{Stage I -- Appearance Training:}
\BlankLine
Add finetuning parameters $\pi_s$ to $\epsilon_\theta$ to obtain $\epsilon_{\theta, \pi_s}$;
\BlankLine
\For{each epoch}{
    \For{each image in image data loader}{
        \BlankLine 
        Encode image to latent space $z_0$ using $vae$;
        \BlankLine 
        Sample noise $\epsilon$ and add to $z_0$ to obtain $z_t$;
        \BlankLine 
        Encode text embeddings $\tau_\theta(y)$;
        \BlankLine
        Compute diffusion loss with predicted noise:
        \[
        \mathcal{L}_{z_0, y, \epsilon_i, t} \left[ \epsilon, \epsilon_{\theta,\pi_s}(z_t, t, \tau_\theta(y)) \right];
        \]
        Update spatial tuning parameters $\pi_s$;
    }
}
\BlankLine 
Save trained spatial tuning checkpoints.
\BlankLine
\textbf{Stage II -- Motion Training:}
\BlankLine 
Add finetuning parameters $\pi_t$ to $\epsilon_\theta$ to obtain $\epsilon_{\theta, \pi_t}$;
\BlankLine
\For{each epoch}{
    \For{each video in video data loader}{
        \BlankLine 
        Encode video to latent space $z_0$ using $vae$;
        \BlankLine 
        Sample noise $\epsilon$ and add to $z_0$ to obtain $z_t$;
        \BlankLine 
        Encode text embeddings $\tau_\theta(y)$;
        \BlankLine
        Compute diffusion loss with predicted noise:
        \[
        \mathcal{L}_{z_0, y, \epsilon_v, t} \left[ \epsilon, \epsilon_{\theta,\pi_t}(z_t, t, \tau_\theta(y)) \right];
        \]
        Update temporal tuning parameters $\pi_t$;
    }
}
\BlankLine 
Save trained temporal tuning checkpoints;
\BlankLine
\textbf{Combined Inference:}
\BlankLine 
Initialize pipeline with pretrained models;
\BlankLine 
Load trained spatial finetuning parameters $\pi_s$;
\BlankLine 
Load trained temporal finetuning parameters $\pi_t$;
\BlankLine 
\For{each prompt}{
    Prepare input latents (random noise) $\hat{z}_t$;
    \BlankLine
    Perform the denoising process step by step, ultimately yielding the denoised latent $\hat{z}_0$:
    \[
    \hat{z}_{t-1} = \hat{z}_t - \hat{\epsilon}_{\theta, \pi_s, \pi_t}(\hat{z}_t, t, \tau_\theta(y));
    \]
    \BlankLine
    Decode denoised latent $z_0$ into video using $vae$;
}
\caption{\textbf{Stage-wise Training Method}}
\label{alg:two-stage_cvg}
\end{algorithm}
\endgroup

A fundamental limitation of the stage-wise strategy lies in its decoupling of appearance and motion optimization into disjoint phases. Because spatial and temporal LoRA parameters are learned independently, their respective adaptations may become misaligned or even contradictory. When jointly applied during inference, such incompatibilities can lead to concept interference. For instance, the appearance-adapted module may emphasize fine-grained spatial details, whereas the motion-adapted module may inadvertently modify spatial structures to satisfy temporal consistency. This conceptual conflict often compromises subject fidelity, resulting in incomplete or inconsistent motion patterns.

JointTuner addresses this challenge through a joint optimization framework that trains a single set of parameters under simultaneous supervision from both appearance (derived from static images) and motion (extracted from videos). By alternately optimizing appearance-focused and motion-focused reconstruction objectives, the gating regulators within GLoRA learn to selectively activate or suppress their associated modules based on whether they specialize in modeling appearance or motion. During inference, this gating mechanism facilitates the dynamic composition of specialized GLoRAs, enabling adaptive integration of appearance and motion features. This joint training paradigm effectively mitigates inter-parameter conflicts and ensures coherent alignment between the customized subject identity and its intended motion, thereby producing more reliable and temporally consistent video generation.

\begin{table*}[t]
    \centering
    \resizebox{\linewidth}{!}{
    \begin{tabular}{l|l|l}
    \hline
        \textbf{Category} & \textbf{Description} & \textbf{Samples} \\ \hline
        \multirow{5}{*}{Rigid Subject} & \multirow{5}{*}{Rigid objects with no articulation or deformable parts.} & guitar (CustomDiffusion/instrument\_music2),   \\ 
        ~ & ~ & book (CustomDiffusion/things\_book), \\ 
        ~ & ~ & car (CustomDiffusion/transport\_car6), \\ 
        ~ & ~ & backpack (DreamBooth/backpack\_dog), \\
        ~ & ~ & clock (DreamBooth/clock) \\ \midrule
        \multirow{5}{*}{Non-Humanoid Subject} & \multirow{5}{*}{Subjects lacking human-like limb structure.} & cat (CustomDiffusion/pet\_cat5),     \\ 
        ~ & ~ & wolf\_plushie (DreamBooth/wolf\_plushie), \\
        ~ & ~ & tortoise\_plushie (CustomDiffusion/plushie\_tortoise), \\ 
        ~ & ~ & unicorn\_toy (CustomDiffusion/toy\_unicorn), \\ 
        ~ & ~ & dog (DreamBooth/dog) \\ \midrule
        \multirow{5}{*}{Humanoid Subject} & \multirow{5}{*}{Subjects with articulated limbs capable of standing posture.} & pink\_plushie (CustomDiffusion/plushie\_pink),     \\ 
        ~ & ~ & bear\_plushie (DreamBooth/bear\_plushie), \\ 
        ~ & ~ & sloth\_plushie (DreamBooth/grey\_sloth\_plushie), \\ 
        ~ & ~ & monster\_toy (DreamBooth/monster\_toy), \\ 
        ~ & ~ & terracotta\_warrior (MotionDirector/Terracotta\_Warrior) \\ \hline \midrule
        \multirow{3}{*}{Rigid Motion} & \multirow{3}{*}{Global movement of rigid objects without internal deformation.} & boat\_sailing (DAVIS2016/boat),   \\ 
        ~ & ~ & bus\_traveling (DAVIS2016/bus), \\ 
        ~ & ~ & train\_turning (DAVIS2016/train) \\ \midrule
        \multirow{3}{*}{Non-Human Motion} & \multirow{3}{*}{Motion patterns typical of non-human entities.} & bear\_walking (DAVIS2016/bear),  \\ 
        ~ & ~ & duck\_walking (DAVIS2016/mallard-fly),  \\ 
        ~ & ~ & dog\_walking (DAVIS2016/dog) \\ \midrule
        \multirow{3}{*}{Human Motion (Without Props)} & \multirow{3}{*}{Body-only motion without interaction with external objects.} & person\_dancing (DAVIS2016/breakdance-flare),   \\ 
        ~ & ~ & person\_twirling (DAVIS2016/dance-twirl), \\ 
        ~ & ~ & person\_walking (DAVIS2016/kite-walk) \\ \midrule
        \multirow{3}{*}{Human Motion (With Props)} & \multirow{3}{*}{Motion involving interaction with props (5 clips per motion).} & person\_lifting\_barbell (UCF101/CleanAndJerk),   \\ 
        ~ & ~ & person\_playing\_cello (UCF101/PlayingCello), \\ 
        ~ & ~ & person\_playing\_flute (UCF101/PlayingFlute) \\ \bottomrule
    \end{tabular}
    }
    \vspace{-1em}
    \caption{Overview of the subject and motion categories in our benchmark dataset. Samples are annotated with their instance names and respective dataset/model origins.}
    \label{tab:dataset_details}
    \vspace{-1em}
\end{table*}

\begin{table}[t]
    \centering
    \resizebox{\linewidth}{!}{
    \begin{tabular}{l|l}
    \hline
        \textbf{Subject Category} & \textbf{Compatible Motion Categories} \\ \hline
        Rigid Subject & Rigid Motion \\ \hline
        \multirow{3}{*}{Non-Humanoid Subject} & Non-Human Motion \\ 
        ~ & Human Motion (Without Props) \\ 
        ~ & Human Motion (With Props) \\ \hline
        \multirow{2}{*}{Humanoid Subject} & Human Motion (Without Props) \\ 
        ~ & Human Motion (With Props) \\ \hline
    \end{tabular}
    }
    \vspace{-1em}
    \caption{Allowed subject-motion category combinations used for generating customized videos.}
    \label{tab:dataset_pairing_info}
    \vspace{-1em}
\end{table}

\section{Benchmark Details}
\label{sec:benchmark}

\subsection{Datasets}

We construct a comprehensive dataset comprising 90 appearance-motion combinations using public sources~\cite{dreambooth, customdiffusion, motiondirector, davis2016, ucf101}. As summarized in Tab.\ref{tab:dataset_details}, the dataset integrates 15 subjects from CustomDiffusion~\cite{customdiffusion}, DreamBooth~\cite{dreambooth}, and MotionDirector~\cite{motiondirector}, along with nine motion types from DAVIS2016~\cite{davis2016} and three composite motions (each comprising 5 video clips) from UCF101~\cite{ucf101}. We design six distinct subject-motion pairing configurations, as shown in Tab.~\ref{tab:dataset_pairing_info}. To further assess generalization, we adopt five diverse background settings, including natural scenes (grassland, jungle, snowscape, and beach) and an urban environment (cobblestone street). This results in a comprehensive benchmark dataset of 450 customized videos, covering diverse combinations of subjects, motions, and backgrounds.

To standardize the motion input format for different backbones, we employ two preprocessing strategies: a 16-frame sequence for UNet-based methods and a 49-frame sequence for DiT-based methods. For UNet-based methods pre-trained on short video segments, each video is first processed to extract a 16-frame sequence. Specifically, for the DAVIS2016 Dataset~\cite {davis2016}, frames are sampled every two frames, whereas for the UCF101 Dataset~\cite {ucf101}, frames are sampled every six frames. This results in 16-frame clips with a frame rate of 8 FPS and a duration of 2 seconds, while maintaining the original video resolution. In addition to the 16-frame setting, we create a 49-frame input setting by directly extracting the first 49 frames from each original video, without temporal downsampling. These longer clips are intended for DiT-based models, such as CogVideoX~\cite{2022cogvideo}, which are pre-trained on longer video sequences and benefit from the extended temporal context. All 49-frame videos are preprocessed to a fixed resolution of $720 \times 480$.

Furthermore, we annotate each video clip using BLIP-2~\cite{blip2} to generate descriptive captions that serve as textual prompts during training. To further improve the quality of prompts for DiT-based methods, we employ Qwen3~\cite{qwen3} to enrich the captions. For example, a simple caption such as ``\textit{A bear plushie playing the flute on the grass}'' is expanded into a richer and more vivid prompt: ``\textit{A bear plushie sitting on a lush green meadow, serenely playing a wooden flute under a gentle, golden sunset}''.

\begin{table*}[t]
    \centering
    \resizebox{\linewidth}{!}{
        \begin{tabular}{c|ccc|ccc|cc|cc|ccc}
            \hline
            \toprule
            \textbf{Models} & \multicolumn{3}{c|}{\underline{\textbf{\textit{Composite Metrics}}}} & \multicolumn{3}{c|}{\underline{\textbf{\textit{Semantic Alignment}}}} & \multicolumn{2}{c|}{\underline{\textbf{\textit{Motion Dynamism}}}} & \multicolumn{2}{c|}{\underline{\textbf{\textit{Temporal Consistency}}}} & \multicolumn{3}{c}{\underline{\textbf{\textit{Perceptual Quality}}}}  \\ 
            \textbf{(on ModelScope~\cite{modelscope})} & \textbf{ARS} $\downarrow$ & \textbf{NAS} $\uparrow$ & \textbf{AAS} $\uparrow$ & \textbf{CLIP-T} $\uparrow$ & \textbf{CLIP-I} $\uparrow$ & \textbf{Mot-Fid} $\uparrow$ & \textbf{Mot-Smth} $\uparrow$ & \textbf{Dyn-Deg} $\uparrow$ & \textbf{Subj-Con} $\uparrow$ & \textbf{Bkgd-Con} $\uparrow$ & \textbf{Pick-Sc} $\uparrow$ & \textbf{Aesth-Q} $\uparrow$ & \textbf{Img-Q} $\uparrow$ \\ \midrule
        \textbf{MotionDirector-MS} & \colorbox{secondpurple}{2.20}  & 92.98  & 61.81  & \colorbox{secondpurple}{28.91}  & \colorbox{bestred}{68.00}  & 67.42  & 96.01  & 37.56  & \colorbox{bestred}{92.40}  & 93.41  & \colorbox{secondpurple}{20.34}  & \colorbox{secondpurple}{51.04}  & \colorbox{secondpurple}{63.05}  \\ 
        \textbf{DreamVideo-MS} & 2.30  & \colorbox{secondpurple}{93.76}  & \colorbox{secondpurple}{62.31}  & 27.69  & 54.44  & \colorbox{bestred}{78.44}  & \colorbox{secondpurple}{96.16}  & \colorbox{bestred}{51.11}  & 91.22  & \colorbox{bestred}{94.85}  & 20.10  & 48.69  & 60.41  \\ \midrule
        \textbf{JointTuner-MS (Ours)} & \colorbox{bestred}{1.50}  & \colorbox{bestred}{97.42}  & \colorbox{bestred}{63.88}  & \colorbox{bestred}{32.08}  & \colorbox{secondpurple}{63.31}  & \colorbox{secondpurple}{69.65}  & \colorbox{bestred}{97.50}  & \colorbox{secondpurple}{48.44}  & \colorbox{secondpurple}{91.47}  & \colorbox{secondpurple}{93.43}  & \colorbox{bestred}{21.10}  & \colorbox{bestred}{56.50}  & \colorbox{bestred}{65.33}  \\ \bottomrule
    \end{tabular}
    }
    \vspace{-1em}
    \caption{Quantitative comparison of UNet-based methods, built on ModelScope. We highlight the \colorbox{bestred}{best} and \colorbox{secondpurple}{second-best} values for each metric.}
    \label{tab:overall_performance_ms}
    \vspace{-1em}
\end{table*}

\subsection{Metrics}

We establish a comprehensive evaluation framework across four dimensions: semantic alignment, motion dynamism, temporal consistency, and perceptual quality. For overall assessment, we report three composite metrics. The specific metrics are described as follows.

\textbf{Composite Metrics}
\begin{itemize}
    \item \textbf{Average Rank Score (ARS):} Mean model ranking score across all ten metrics in the same table.
    \item \textbf{Normalized Average Score (NAS):} Scores are normalized against the top-performing method (scaled to 100) within a single table, and then averaged across all evaluation metrics for each model's performance.
    \item \textbf{Absolute Average Score (AAS):} Arithmetic mean of raw metric scores.
\end{itemize}

\textbf{Semantic Alignment}
\begin{itemize}
    \item \textbf{CLIP-Text (CLIP-T):} Measures text-video alignment via frame-wise cosine similarity between CLIP~\cite{clip} embeddings of input prompts and generated frames.
    \item \textbf{CLIP-Image (CLIP-I):} Evaluates visual-semantic correspondence using CLIP~\cite{clip} image encoder to compare reference images and generated frames.
    \item \textbf{Motion-Fidelity (Mot-Fid):} Assesses motion pattern consistency using CoTracker3~\cite{cotracker3} from diffusion-motion-transfer~\cite{diffusionmotiontransfer}.
\end{itemize}

\textbf{Motion Dynamism}
\begin{itemize}
    \item \textbf{Motion-Smoothness (Mot-Smth):} Evaluates temporal coherence using video interpolation priors from AMT~\cite{amt} via VBench~\cite{vbench}.
    \item \textbf{Dynamic-Degree (Dyn-Deg):} Quantifies motion intensity using optical flow estimation via RAFT~\cite{raft} within VBench~\cite{vbench}.
\end{itemize}

\textbf{Temporal Consistency}
\begin{itemize}
    \item \textbf{Subject-Consistency (Subj-Con):} Tracks object persistence using DINO~\cite{dino} feature similarity via VBench~\cite{vbench}.
    \item \textbf{Background-Consistency (Bkgd-Con):} Evaluates background stability using CLIP~\cite{clip} feature similarity across frames via VBench~\cite{vbench}.
\end{itemize}

\textbf{Perceptual Quality}
\begin{itemize}
    \item \textbf{Pick-Score (Pick-Sc):} Predicts human preference scores using PickScore~\cite{pickscore} with frame-level averaging.
    \item \textbf{Aesthetic-Quality (Aesth-Q):} Measures artistic merit using LAION aesthetic predictor~\cite{aestheticpredictor} via VBench~\cite{vbench}.
    \item \textbf{Imaging-Quality (Img-Q):} Evaluates technical quality using MUSIQ~\cite{musiq} trained on SPAQ~\cite{spaq} via VBench~\cite{vbench}.
\end{itemize}

\section{Experiment Settings}
\label{sec:exp_setting}

\subsection{Implementation Details}
Our JointTuner is compatible with both UNet-based and DiT-based diffusion models. The implementation details are provided below.
\begin{itemize}
\item \textbf{UNet-based JointTuner.} The joint training framework uses GLoRA parameters in transformers of UNet, including ZeroScope~\cite{zeroscope} and ModelScope~\cite{modelscope}, trained for 2000 iterations with a learning rate of $5 \times 10^{-4}$. Training requires 17GB of GPU memory and takes approximately 30 minutes on an RTX 4090. Inference utilizes DDIM~\cite{ddim} with 30 sampling steps to generate 16-frame videos at 8 FPS with a resolution of $384 \times 384$.

\item \textbf{DiT-based JointTuner.} The joint training framework uses GLoRA parameters in transformers of DiT, including CogVideoX~\cite{2022cogvideo}, trained for 1000 iterations with a learning rate of $1 \times 10^{-4}$. Training requires 20GB of GPU memory for CogVideoX-2B and 38GB for CogVideoX-5B. The training duration is approximately 60 minutes for a task involving five images and a single clip video, and around 120 minutes for a task with five images and five clip videos, based on the A100 (40 GB). Inference is performed with 50 sampling steps to generate 49-frame videos at 8 FPS and a resolution of $720 \times 480$.
\end{itemize}

\subsection{Comparison Methods}
We compare JointTuner with several open-source methods under consistent evaluation settings without external controls. For stage-wise training, we use ZeroScope~\cite{zeroscope} and ModelScope~\cite{modelscope} as base models and compare against \textbf{MotionDirector}~\cite{motiondirector} and \textbf{DreamVideo}~\cite{dreamvideo}. For joint training, CogVideoX~\cite{2022cogvideo} serves as the baseline. We further compare JointTuner (implemented on both CogVideoX-2B and CogVideoX-5B) with \textbf{CogVideo-5B}, its \textbf{LoRA} variant, and \textbf{DualReal}~\cite{dualreal}. Below is an overview of the state-of-the-art methods along with the experimental setup.

\begin{itemize}
    \item \textbf{MotionDirector~\cite{motiondirector}}: MotionDirector decouples appearance and motion learning via dual-path LoRA modules and introduces an appearance-debiased temporal loss to improve motion generalization across diverse subjects. We use the settings from the official MotionDirector implementation. For appearance learning, the spatial LoRA is trained for 1,000 iterations at a learning rate of $5.0 \times 10^{-4}$. For motion learning, the temporal LoRA is trained for 3,000 iterations at $5.0 \times 10^{-4}$ for multi-reference motions and 300 iterations for single-reference motions. During inference, 16-frame videos at 8 FPS are generated using DDIM with 30 sampling steps.
    \item \textbf{DreamVideo~\cite{dreamvideo}}: DreamVideo separates training into subject and motion stages, using identity and motion adapters with appearance-guided motion learning to enable flexible subject-motion composition. We follow the settings from the official DreamVideo implementation. For appearance learning, the text identity is optimized for 3,000 iterations at $1.0 \times 10^{-4}$, and the identity adapter is trained for 1,000 iterations at $1.0 \times 10^{-5}$. For motion learning, the motion adapter is trained for 1,000 iterations at $1.0 \times 10^{-5}$ for multi-reference motions and 3,000 iterations for single-reference motions. During inference, 16-frame videos at 8 FPS are generated using DDIM with 50 sampling steps and classifier-free guidance.
    \item \textbf{DualReal~\cite{dualreal}}: DualReal employs joint training with alternating optimization phases and a controller to achieve lossless fusion of identity and motion without cross-dimensional conflict. We follow the settings from the official implementation of DualReal. DualReal is trained for 1000 iterations with a learning rate of $1 \times 10^{-3}$ based on CogVideoX-5B~\cite{2022cogvideo}. During inference, 49-frame videos at 8 FPS are generated with 50 sampling steps.
\end{itemize}

\section{Extended Analysis and Additional Results}
\label{sec:more_results}


\subsection{Overall comparison on ModelScope}

We conducted experiments on another UNet-based model, ModelScope~\cite{modelscope}, and the results are shown in Tab.~\ref{tab:overall_performance_ms}. JointTuner leads in perceptual quality and consistently ranks in the top-2 across most metrics. While MotionDirector excels in static fidelity (high CLIP-I, Subj-Con), it falls short in motion capture. On the other hand, DreamVideo captures motion well (high CLIP-T, Mot-Fid) but suffers from degradation in appearance. In contrast, JointTuner generates videos with better overall appearance and motion.



\begin{figure}[t]
  \centering
  \includegraphics[width=\linewidth]{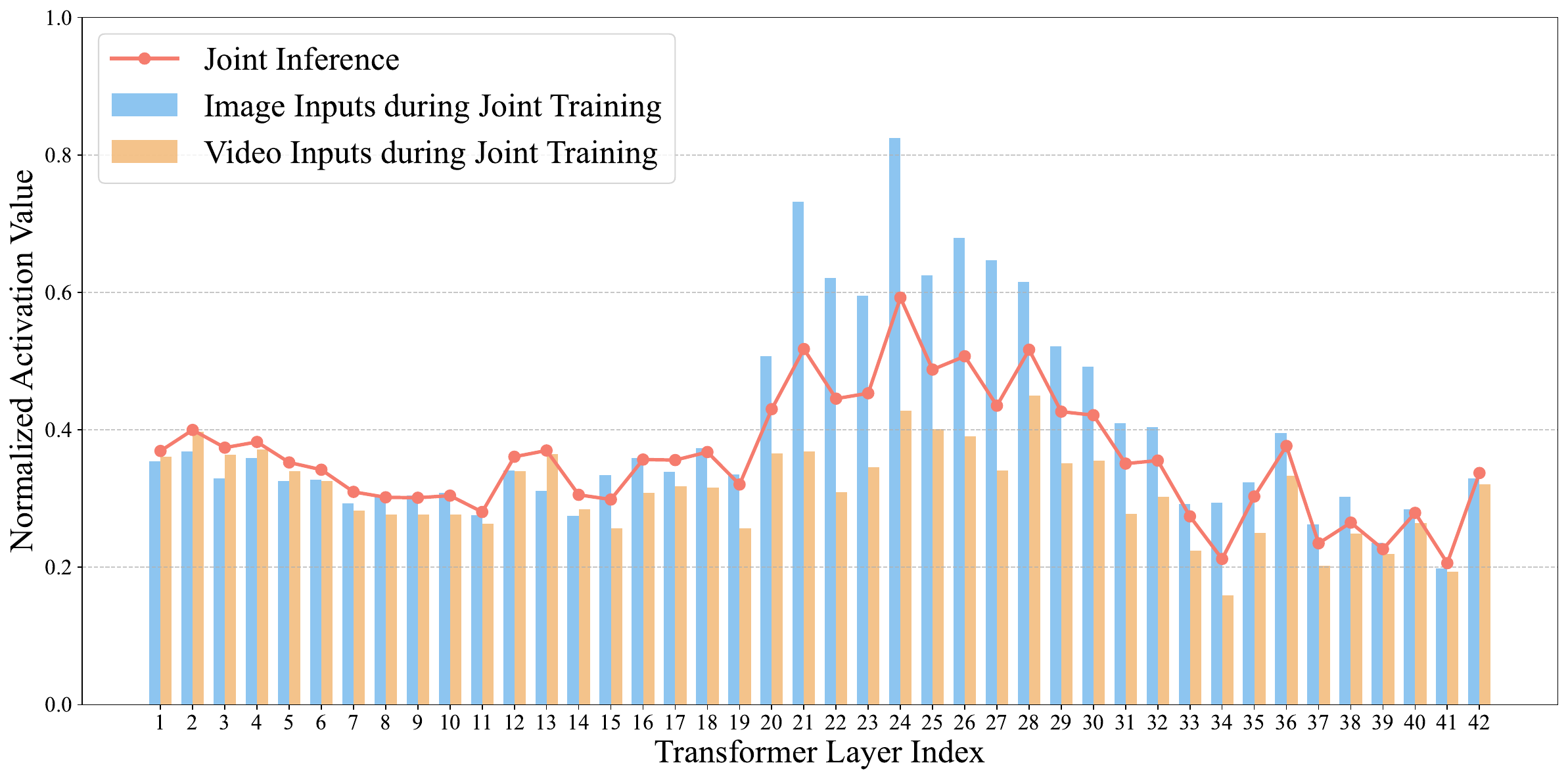}
  \vspace{-2em}
  \caption{Activation values of gating regulators across different Transformer layers in CogVideoX-5B during both training and inference.}
  \label{fig:regulation_values_cog}
  \vspace{-1em}
\end{figure}

\begin{figure}[t]
  \centering
  \includegraphics[width=\linewidth]{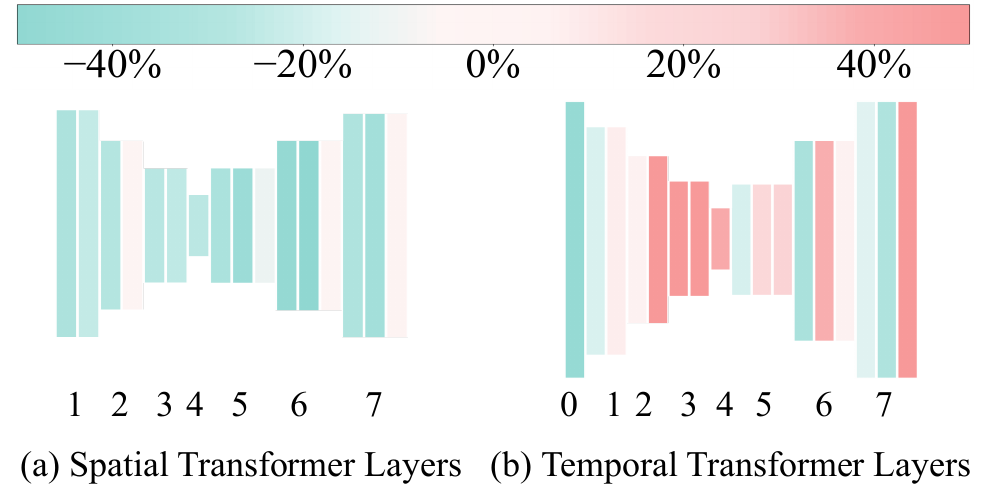}
  \vspace{-2em}
  \caption{Changes in the values of gating regulators across different Transformer layers when transitioning from image input to video input, as observed in ZeroScope.}
  \label{fig:regulation_values_zs}
  \vspace{-1em}
\end{figure}

\subsection{Analysis of Gated Modulation in GLoRA}

This section presents a comprehensive analysis of how GLoRA’s gated modulation mechanism enables adaptive feature learning during joint training and feature fusion during joint inference. We select 30 appearance-motion pairs for the experiment.

\textbf{Gated Regulator Analysis on CogVideoX-5B.} We record the activation magnitudes of GLoRA across all Transformer layers under three settings: (i) joint training with image inputs, (ii) joint training with video inputs, and (iii) joint inference. The layer-wise activation patterns reveal a clear modality-driven specialization. Specifically, in the early layers (1–14), video inputs mostly elicit stronger activations than image inputs, suggesting a heightened sensitivity to motion-related features. In contrast, in the deeper layers (15–42), image inputs dominate, reflecting a greater emphasis on static appearance features. This stratified behavior indicates that distinct GLoRA modules naturally specialize according to input modality during joint training. Crucially, during inference, the gating mechanism dynamically refines this specialization in a context-aware manner. In motion-sensitive early layers, the unified activation values often exceed those observed during either image-only or video-only training, implying an intentional amplification of motion signals at test time. Conversely, in appearance-biased later layers, inference activations typically lie between the image- and video-only training values, frequently approximating their average, suggesting a deliberate attenuation of overly dominant appearance responses. Through these complementary adjustments, GLoRA achieves a balanced and adaptive composition of modality-specific experts, suppressing potentially conflicting ones, thereby enabling coherent joint modeling of appearance and motion and effectively mitigating concept interference.

\textbf{Gated Regulator Analysis on ZeroScope.} Leveraging the UNet architecture’s explicit separation of spatial and temporal Transformer blocks, we directly monitor GLoRA activation patterns when switching the input modality from images to videos during joint training. As shown in Fig.~\ref{fig:regulation_values_zs}, the activations in spatial Transformer layers drop significantly upon transitioning to video inputs, while those in temporal Transformer layers rise markedly. This shift confirms that GLoRA employs learnable modules that dynamically modulate the contributions of spatial and temporal LoRA pathways based on the input modality.


\begin{table*}[t]
    \centering
    \resizebox{\linewidth}{!}{
        \begin{tabular}{c|ccc|ccc|cc|cc|ccc}
            \hline
            \toprule
            \textbf{Models} & \multicolumn{3}{c|}{\underline{\textbf{\textit{Composite Metrics}}}} & \multicolumn{3}{c|}{\underline{\textbf{\textit{Semantic Alignment}}}} & \multicolumn{2}{c|}{\underline{\textbf{\textit{Motion Dynamism}}}} & \multicolumn{2}{c|}{\underline{\textbf{\textit{Temporal Consistency}}}} & \multicolumn{3}{c}{\underline{\textbf{\textit{Perceptual Quality}}}}  \\ 
            \textbf{(on ZeroScope~\cite{zeroscope})} & \textbf{ARS} $\downarrow$ & \textbf{NAS} $\uparrow$ & \textbf{AAS} $\uparrow$ & \textbf{CLIP-T} $\uparrow$ & \textbf{CLIP-I} $\uparrow$ & \textbf{Mot-Fid} $\uparrow$ & \textbf{Mot-Smth} $\uparrow$ & \textbf{Dyn-Deg} $\uparrow$ & \textbf{Subj-Con} $\uparrow$ & \textbf{Bkgd-Con} $\uparrow$ & \textbf{Pick-Sc} $\uparrow$ & \textbf{Aesth-Q} $\uparrow$ & \textbf{Img-Q} $\uparrow$ \\ \midrule
        \textbf{w/o GLoRA} & 3.60  & 94.26  & 62.49  & 28.94  & 60.18  & 72.03  & 96.20  & 46.89  & 91.14  & 93.31  & 20.46  & 52.23  & 63.54   \\ 
        \textbf{Decoupled LoRA} & 3.20  & 95.20  & 63.05  & 28.95  & 59.89  & \colorbox{secondpurple}{73.83}  & 96.01  & \colorbox{secondpurple}{48.89}  & 91.21  & 93.21  & 20.35  & 54.62  & \colorbox{secondpurple}{63.59}   \\ 
        \textbf{Two-Stage Training} & 3.50  & 92.94  & 62.26  & 28.37  & \colorbox{bestred}{69.08}  & 73.24  & 95.54  & 34.67  & \colorbox{bestred}{93.11}  & \colorbox{bestred}{94.01}  & 20.29  & 51.57  & 62.68   \\ 
        \textbf{w/o AiT Loss} & \colorbox{secondpurple}{2.80}  & \colorbox{secondpurple}{96.53}  & \colorbox{secondpurple}{63.86}  & \colorbox{secondpurple}{29.68}  & 60.30  & \colorbox{bestred}{80.08}  & \colorbox{secondpurple}{96.71}  & \colorbox{bestred}{49.33}  & 90.82  & 93.11  & \colorbox{secondpurple}{20.83}  & \colorbox{secondpurple}{55.04}  & 62.64   \\ \midrule
        \textbf{JointTuner-ZS (Ours)} & \colorbox{bestred}{1.80}  & \colorbox{bestred}{97.46}  & \colorbox{bestred}{64.15}  & \colorbox{bestred}{32.14}  & \colorbox{secondpurple}{62.44}  & 72.61  & \colorbox{bestred}{97.78}  & 46.89  & \colorbox{secondpurple}{91.92}  & \colorbox{secondpurple}{93.83}  & \colorbox{bestred}{21.19}  & \colorbox{bestred}{57.14}  & \colorbox{bestred}{65.53}   \\ \bottomrule
    \end{tabular}
    }
    \caption{Quantitative ablation results for each core component of the proposed JointTuner, on ZeroScope. We highlight the \colorbox{bestred}{best} and \colorbox{secondpurple}{second-best} values for each metric.}
    \label{tab:ablation}
    \vspace{-10pt}
\end{table*}

\begin{figure}[t]
  \centering
  \includegraphics[width=\linewidth]{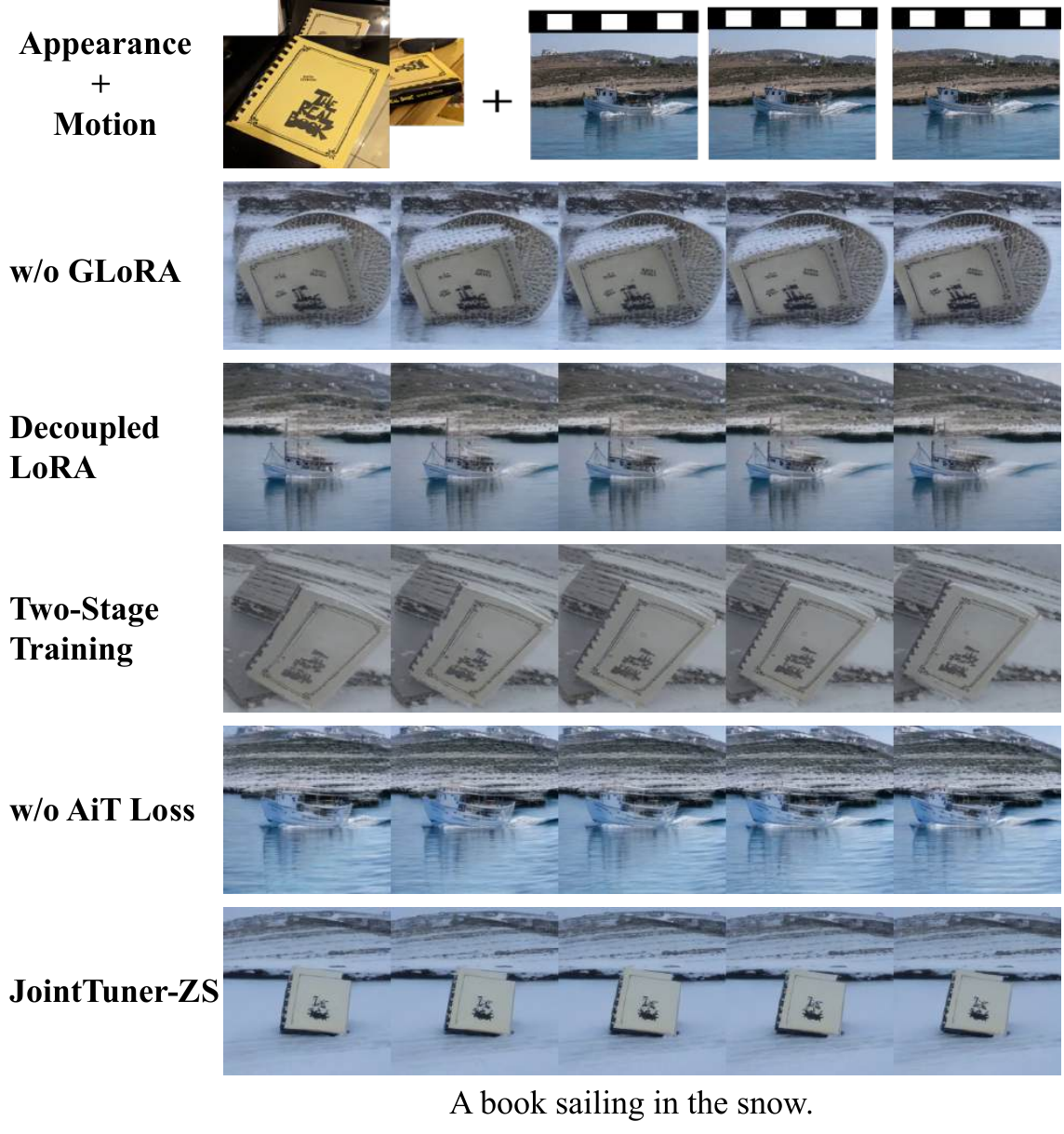}
  \vspace{-1em}
  \caption{Qualitative ablation results for each component of JointTuner based on ZeroScope.}
  \label{fig:ablation_sample}
  \vspace{-1em}
\end{figure}

\subsection{Ablation Studies on ZeroScope}

We further conduct ablation studies on the UNet-based model ZeroScope, evaluating four key variants: 
\textbf{(\romannumeral 1)~w/o GLoRA}, which employs standard LoRA without gating regulation; 
\textbf{(\romannumeral 2)~Decoupled LoRA}, which alternately optimizes spatial and temporal typical LoRA modules; 
\textbf{(\romannumeral 3)~Two-Stage Training}, which separates appearance and motion learning into distinct training phases; and 
\textbf{(\romannumeral 4)~w/o AiT Loss}, which uses the original diffusion loss instead of AiT loss when processing video inputs during joint training.

\textbf{Quantitative results.}
As shown in Tab.~\ref{tab:ablation}, JointTuner-ZS significantly outperforms the \textit{w/o GLoRA} variant, reducing ARS from 3.6 to 1.8 and improving NAS (+3.2) and AAS (+1.66). Moreover, \textit{w/o GLoRA} performs worse than \textit{Decoupled LoRA}, further demonstrating that gating regulation enhances the effectiveness of unified LoRA. JointTuner-ZS also surpasses both \textit{Decoupled LoRA} and \textit{Two-Stage Training}. Specifically, \textit{Decoupled LoRA} yields degraded CLIP-T and CLIP-I scores, while \textit{Two-Stage Training} further reduces perceptual quality across Pick-Sc, Aesth-Q, and Img-Q. These results highlight the superiority of joint optimization in aligning spatial and temporal features. Additionally, \textit{w/o AiT Loss} exhibits notable drops in CLIP-T and CLIP-I, indicating that the AiT loss is crucial for disentangling motion from appearance while preserving the target subject’s visual identity.

\textbf{Qualitative results.}
As shown in Fig.~\ref{fig:ablation_sample}, removing gating regulation (\textit{w/o GLoRA}) causes motion artifacts and appearance degradation. The \textit{Decoupled LoRA} variant produces incomplete appearance reconstruction, while \textit{Two-Stage Training} further compromises motion coherence. In contrast, \textit{w/o AiT Loss} suffers from appearance corruption, underscoring the importance of the AiT loss in maintaining visual fidelity.

\subsection{Hyperparameter Sensitivity}
\begin{figure}[t]
    \centering
    \includegraphics[width=\linewidth]{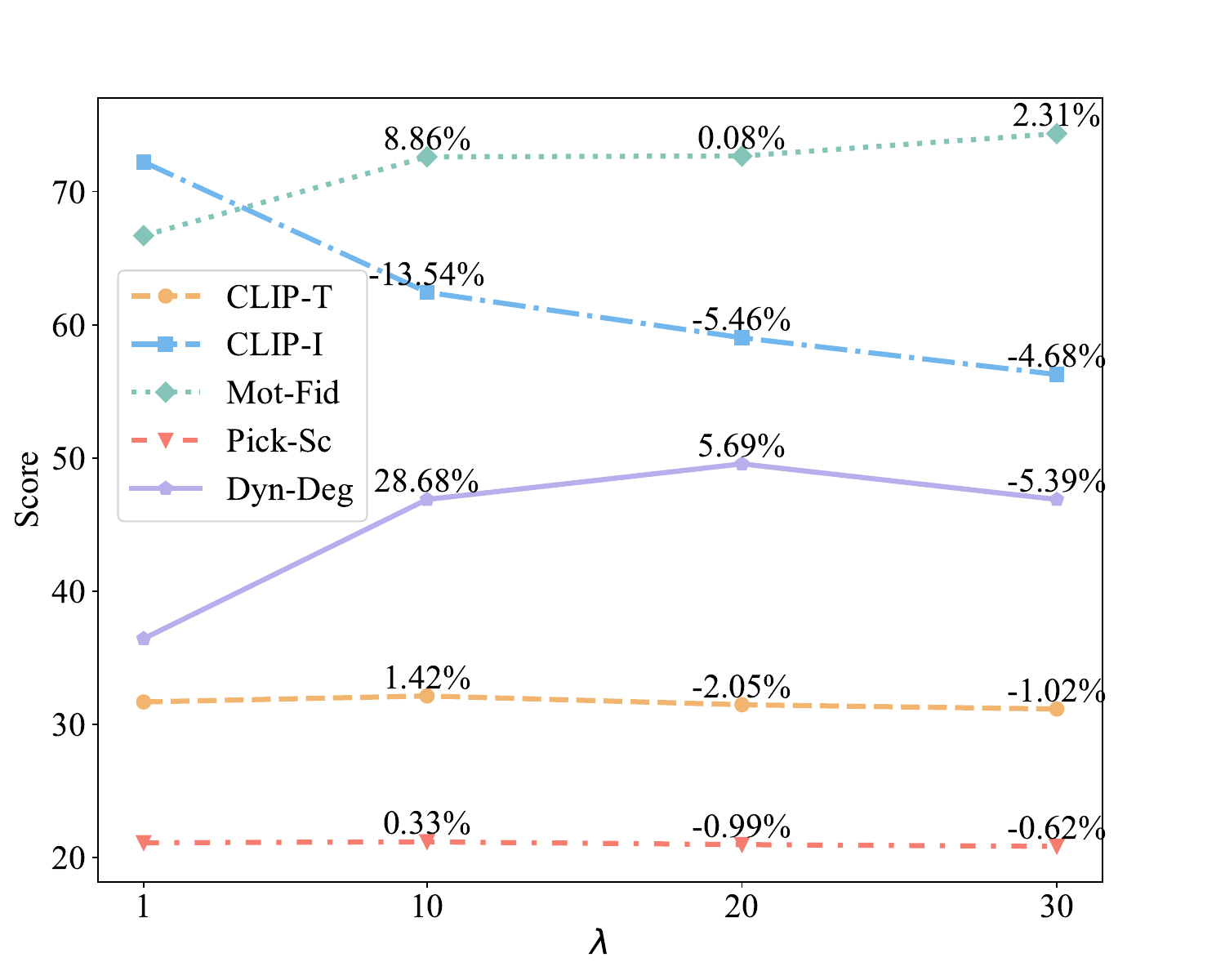}
    \vspace{-2em}
    \caption{Quantitative analysis of the performance of our JointTuner based on ZeroScope, with varying AiT loss weight, $\lambda$.}
    \label{fig:lambda_changes_result}
    \vspace{-1em}
\end{figure}

Since the loss functions for appearance learning and motion learning differ, we also experimented with adjusting the weight of the AiT Loss, denoted by $\lambda$, to evaluate its effect on motion quality in customized video generation. We tested various values of $\lambda$ and analyzed five key metrics, with results presented in Fig.~\ref{fig:lambda_changes_result}. Notably, Dyn-Deg and Mot-Fid improve by 28.68\% and 8.86\%, respectively, when $\lambda=10$, while CLIP-I decreases by 13.54\%. In contrast, CLIP-T and Pick-Sc exhibit minimal variation, indicating lower sensitivity. While $\lambda=10$ provides the optimal performance, the results also show that using $\lambda=1$ does not lead to significant degradation. In practice, if a setting with $\lambda=1$ produces unsatisfactory results, increasing it to $\lambda=10$ can often yield better motion quality, making $\lambda=10$ a reliable reference for future customizations.

\subsection{More Customized Videos}

In this section, we showcase more appearance-motion combined customized videos generated by JointTuner and provide a qualitative comparison with current state-of-the-art methods. The results are as follows.

\begin{figure*}[t]
  \centering
    \includegraphics[width=\linewidth]{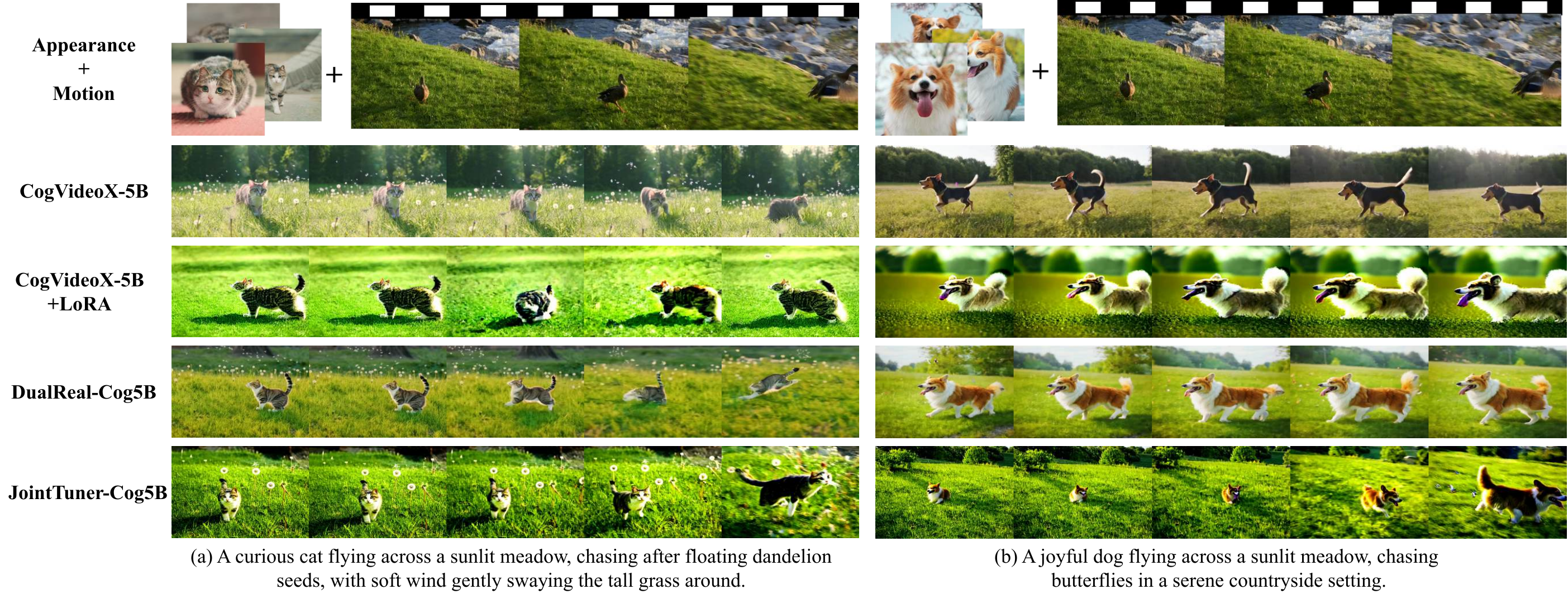}
    \vspace{-2em}
    \caption{Qualitative results of appearance-motion combined customization on CogVideoX-5B.}
    \label{fig:more_result_cog5b_2}
    \vspace{-1em}
\end{figure*}

\begin{figure*}[t]
  \centering
    \includegraphics[width=\linewidth]{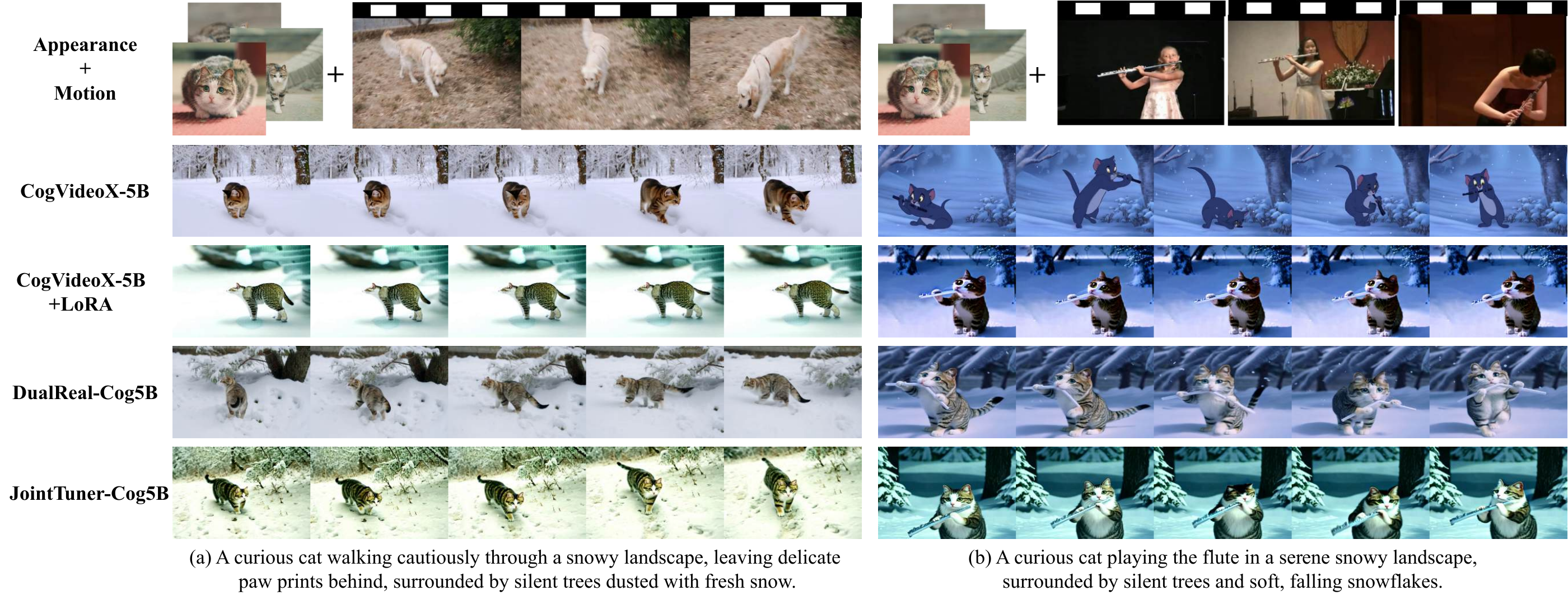}
    \vspace{-2em}
    \caption{Qualitative results of appearance-motion combined customization on CogVideoX-5B.}
    \label{fig:more_result_cog5b_3}
    \vspace{-1em}
\end{figure*}

\begin{figure*}[t]
  \centering
    \includegraphics[width=\linewidth]{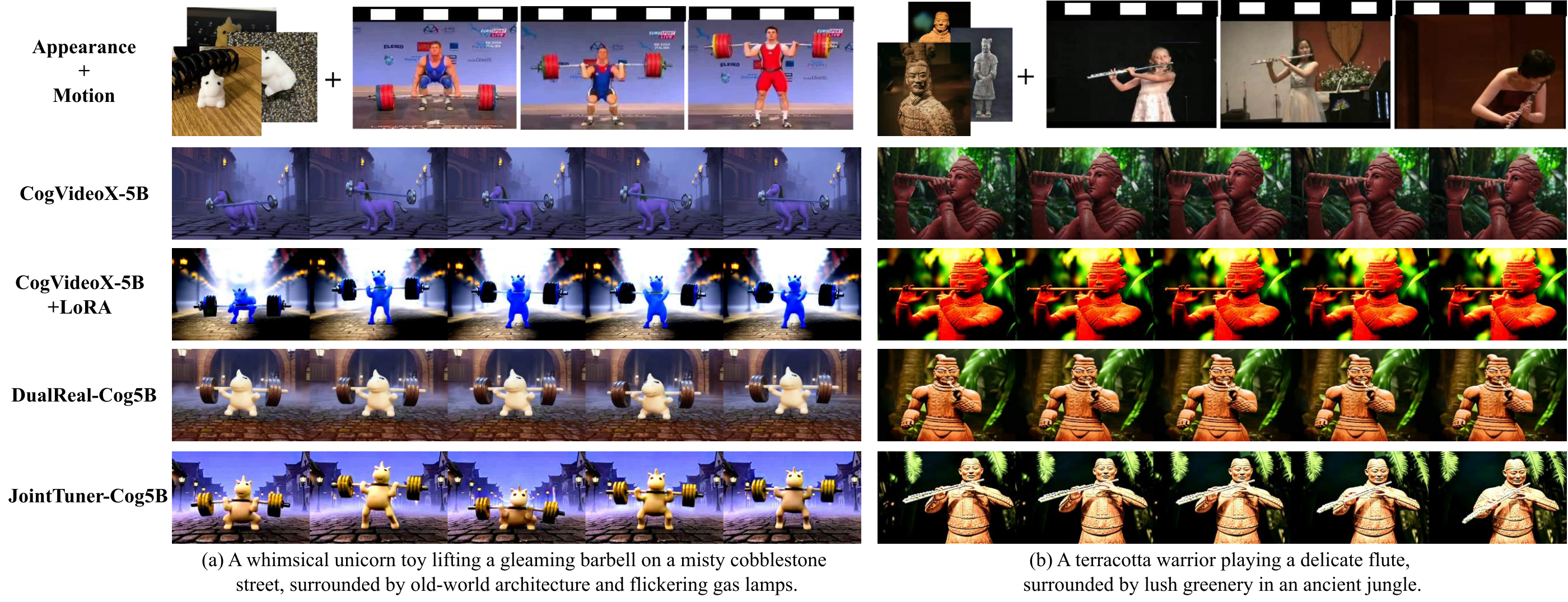}
    \vspace{-2em}
    \caption{Qualitative results of appearance-motion combined customization on CogVideoX-5B.}
    \label{fig:more_result_cog5b_4}
    \vspace{-1em}
\end{figure*}

%% file: main.bib
@inproceedings{dit,
  title={Scalable diffusion models with transformers},
  author={Peebles, William and Xie, Saining},
  booktitle={Proceedings of the IEEE/CVF international conference on computer vision},
  pages={4195--4205},
  year={2023}
}

@article{animatediff,
  title={AnimateDiff: Animate Your Personalized Text-to-Image Diffusion Models without Specific Tuning},
  author={Guo, Yuwei and Yang, Ceyuan and Rao, Anyi and Liang, Zhengyang and Wang, Yaohui and Qiao, Yu and Agrawala, Maneesh and Lin, Dahua and Dai, Bo},
  journal={International Conference on Learning Representations},
  year={2024}
}

@article{modelscope,
  title={Modelscope text-to-video technical report},
  author={Wang, Jiuniu and Yuan, Hangjie and Chen, Dayou and Zhang, Yingya and Wang, Xiang and Zhang, Shiwei},
  journal={arXiv preprint arXiv:2308.06571},
  year={2023}
}

@article{stablevideodiffusion,
  title={Stable video diffusion: Scaling latent video diffusion models to large datasets},
  author={Blattmann, Andreas and Dockhorn, Tim and Kulal, Sumith and Mendelevitch, Daniel and Kilian, Maciej and Lorenz, Dominik and Levi, Yam and English, Zion and Voleti, Vikram and Letts, Adam and others},
  journal={arXiv preprint arXiv:2311.15127},
  year={2023}
}

@inproceedings{blip2,
  title={Blip-2: Bootstrapping language-image pre-training with frozen image encoders and large language models},
  author={Li, Junnan and Li, Dongxu and Savarese, Silvio and Hoi, Steven},
  booktitle={International conference on machine learning},
  pages={19730--19742},
  year={2023},
  organization={PMLR}
}

@online{zeroscope,
    title = {ZeroScope},
    author = {Spencer Sterling},
    year = {2023},
    note = {https://huggingface.co/cerspense/zeroscope\_v2\_576w}
}

@inproceedings{clip,
  title={Learning transferable visual models from natural language supervision},
  author={Radford, Alec and Kim, Jong Wook and Hallacy, Chris and Ramesh, Aditya and Goh, Gabriel and Agarwal, Sandhini and Sastry, Girish and Askell, Amanda and Mishkin, Pamela and Clark, Jack and others},
  booktitle={International conference on machine learning},
  pages={8748--8763},
  year={2021},
  organization={PMLR}
}

@inproceedings{cotracker3,
  title     = {CoTracker3: Simpler and Better Point Tracking by Pseudo-Labelling Real Videos},
  author    = {Nikita Karaev and Iurii Makarov and Jianyuan Wang and Natalia Neverova and Andrea Vedaldi and Christian Rupprecht},
  booktitle = {Proc. {arXiv:2410.11831}},
  year      = {2024}
}

@InProceedings{vbench,
     title={{VBench}: Comprehensive Benchmark Suite for Video Generative Models},
     author={Huang, Ziqi and He, Yinan and Yu, Jiashuo and Zhang, Fan and Si, Chenyang and Jiang, Yuming and Zhang, Yuanhan and Wu, Tianxing and Jin, Qingyang and Chanpaisit, Nattapol and Wang, Yaohui and Chen, Xinyuan and Wang, Limin and Lin, Dahua and Qiao, Yu and Liu, Ziwei},
     booktitle={Proceedings of the IEEE/CVF Conference on Computer Vision and Pattern Recognition},
     year={2024}
 }

@inproceedings{spaq,
  title={Perceptual quality assessment of smartphone photography},
  author={Fang, Yuming and Zhu, Hanwei and Zeng, Yan and Ma, Kede and Wang, Zhou},
  booktitle={Proceedings of the IEEE/CVF conference on computer vision and pattern recognition},
  pages={3677--3686},
  year={2020}
}

@inproceedings{musiq,
  title={Musiq: Multi-scale image quality transformer},
  author={Ke, Junjie and Wang, Qifei and Wang, Yilin and Milanfar, Peyman and Yang, Feng},
  booktitle={Proceedings of the IEEE/CVF international conference on computer vision},
  pages={5148--5157},
  year={2021}
}

@online{aestheticpredictor,
    title = {aesthetic-predictor},
    author = {LAION-AI},
    year = {2022},
    note = {https://github.com/LAION-AI/aesthetic-predictor},
}

@article{ddim,
  title={Denoising diffusion implicit models},
  author={Song, Jiaming and Meng, Chenlin and Ermon, Stefano},
  journal={arXiv preprint arXiv:2010.02502},
  year={2020}
}

@article{pickscore,
  title={Pick-a-pic: An open dataset of user preferences for text-to-image generation},
  author={Kirstain, Yuval and Polyak, Adam and Singer, Uriel and Matiana, Shahbuland and Penna, Joe and Levy, Omer},
  journal={Advances in Neural Information Processing Systems},
  volume={36},
  pages={36652--36663},
  year={2023}
}

@inproceedings{amt,
  title={Amt: All-pairs multi-field transforms for efficient frame interpolation},
  author={Li, Zhen and Zhu, Zuo-Liang and Han, Ling-Hao and Hou, Qibin and Guo, Chun-Le and Cheng, Ming-Ming},
  booktitle={Proceedings of the IEEE/CVF Conference on Computer Vision and Pattern Recognition},
  pages={9801--9810},
  year={2023}
}

@inproceedings{dino,
  title={Emerging properties in self-supervised vision transformers},
  author={Caron, Mathilde and Touvron, Hugo and Misra, Ishan and J{\'e}gou, Herv{\'e} and Mairal, Julien and Bojanowski, Piotr and Joulin, Armand},
  booktitle={Proceedings of the IEEE/CVF international conference on computer vision},
  pages={9650--9660},
  year={2021}
}

@inproceedings{raft,
  title={Raft: Recurrent all-pairs field transforms for optical flow},
  author={Teed, Zachary and Deng, Jia},
  booktitle={Computer Vision--ECCV 2020: 16th European Conference, Glasgow, UK, August 23--28, 2020, Proceedings, Part II 16},
  pages={402--419},
  year={2020},
  organization={Springer}
}

@inproceedings{dreambooth,
  title={Dreambooth: Fine tuning text-to-image diffusion models for subject-driven generation},
  author={Ruiz, Nataniel and Li, Yuanzhen and Jampani, Varun and Pritch, Yael and Rubinstein, Michael and Aberman, Kfir},
  booktitle={Proceedings of the IEEE/CVF conference on computer vision and pattern recognition},
  pages={22500--22510},
  year={2023}
}

@article{customdiffusion,
  title={Multi-Concept Customization of Text-to-Image Diffusion},
  author={Kumari, Nupur and Zhang, Bingliang and Zhang, Richard and Shechtman, Eli and Zhu, Jun-Yan},
  booktitle = {Proceedings of the IEEE/CVF Conference on Computer Vision and Pattern Recognition (CVPR)},
  year      = {2023}
}

@article{ucf101,
  title={UCF101: A dataset of 101 human actions classes from videos in the wild},
  author={Soomro, K},
  journal={arXiv preprint arXiv:1212.0402},
  year={2012}
}

@inproceedings{davis2016,
  author = {F. Perazzi and J. Pont-Tuset and B. McWilliams and L. {Van Gool} and M. Gross and A. Sorkine-Hornung},
  title = {A Benchmark Dataset and Evaluation Methodology for Video Object Segmentation},
  booktitle = {Computer Vision and Pattern Recognition},
  year = {2016}
}

@inproceedings{makeavideo,
  title={Make-a-video: Text-to-video generation without text-video data},
  author={Singer, Uriel and Polyak, Adam and Hayes, Thomas and Yin, Xi and An, Jie and Zhang, Songyang and Hu, Qiyuan and Yang, Harry and Ashual, Oron and Gafni, Oran and others},
  journal={arXiv preprint arXiv:2209.14792},
  year={2022}
}

@inproceedings{videoldm,
  title={Align your latents: High-resolution video synthesis with latent diffusion models},
  author={Blattmann, Andreas and Rombach, Robin and Ling, Huan and Dockhorn, Tim and Kim, Seung Wook and Fidler, Sanja and Kreis, Karsten},
  booktitle={Proceedings of the IEEE/CVF Conference on Computer Vision and Pattern Recognition},
  pages={22563--22575},
  year={2023}
}

@inproceedings{text2videozero,
  title={Text2video-zero: Text-to-image diffusion models are zero-shot video generators},
  author={Khachatryan, Levon and Movsisyan, Andranik and Tadevosyan, Vahram and Henschel, Roberto and Wang, Zhangyang and Navasardyan, Shant and Shi, Humphrey},
  booktitle={Proceedings of the IEEE/CVF International Conference on Computer Vision},
  pages={15954--15964},
  year={2023}
}

@inproceedings{conditionalgan,
  title={Conditional GAN with discriminative filter generation for text-to-video synthesis},
  author={Balaji, Yogesh and Min, Martin Renqiang and Bai, Bing and Chellappa, Rama and Graf, Hans Peter},
  booktitle={Proceedings of the 28th International Joint Conference on Artificial Intelligence},
  pages={1995--2001},
  year={2019}
}

@inproceedings{lvg,
  title={Long video generation with time-agnostic vqgan and time-sensitive transformer},
  author={Ge, Songwei and Hayes, Thomas and Yang, Harry and Yin, Xi and Pang, Guan and Jacobs, David and Huang, Jia-Bin and Parikh, Devi},
  booktitle={European Conference on Computer Vision},
  pages={102--118},
  year={2022},
  organization={Springer}
}

@inproceedings{cogvideo,
  title={Cogvideo: Large-scale pretraining for text-to-video generation via transformers},
  author={Hong, Wenyi and Ding, Ming and Zheng, Wendi and Liu, Xinghan and Tang, Jie},
  journal={arXiv preprint arXiv:2205.15868},
  year={2022}
}

@inproceedings{ccvs,
  title={CCVS: context-aware controllable video synthesis},
  author={Moing, Guillaume Le and Ponce, Jean and Schmid, Cordelia},
  booktitle={Proceedings of the 35th International Conference on Neural Information Processing Systems},
  pages={14042--14055},
  year={2021}
}

@inproceedings{mocogan,
  title={Mocogan: Decomposing motion and content for video generation},
  author={Tulyakov, Sergey and Liu, Ming-Yu and Yang, Xiaodong and Kautz, Jan},
  booktitle={Proceedings of the IEEE conference on computer vision and pattern recognition},
  pages={1526--1535},
  year={2018}
}

@inproceedings{motiondirector,
  title={Motiondirector: Motion customization of text-to-video diffusion models},
  author={Zhao, Rui and Gu, Yuchao and Wu, Jay Zhangjie and Zhang, David Junhao and Liu, Jia-Wei and Wu, Weijia and Keppo, Jussi and Shou, Mike Zheng},
  booktitle={European Conference on Computer Vision},
  pages={273--290},
  year={2024},
  organization={Springer},
  refsection = 1
}

@inproceedings{dreamvideo,
  title={Dreamvideo: Composing your dream videos with customized subject and motion},
  author={Wei, Yujie and Zhang, Shiwei and Qing, Zhiwu and Yuan, Hangjie and Liu, Zhiheng and Liu, Yu and Zhang, Yingya and Zhou, Jingren and Shan, Hongming},
  booktitle={Proceedings of the IEEE/CVF Conference on Computer Vision and Pattern Recognition},
  pages={6537--6549},
  year={2024}
}

@article{customttt,
  title={CustomTTT: Motion and Appearance Customized Video Generation via Test-Time Training},
  author={Bi, Xiuli and Lu, Jian and Liu, Bo and Cun, Xiaodong and Zhang, Yong and Li, Weisheng and Xiao, Bin},
  journal={arXiv preprint arXiv:2412.15646},
  year={2024}
}

@inproceedings{tuneavideo,
  title={Tune-A-Video: One-Shot Tuning of Image Diffusion Models for Text-to-Video Generation},
  author={Wu, Jay Zhangjie and Ge, Yixiao and Wang, Xintao and Lei, Stan Weixian and Gu, Yuchao and Shi, Yufei and Hsu, Wynne and Shan, Ying and Qie, Xiaohu and Shou, Mike Zheng},
  booktitle={2023 IEEE/CVF International Conference on Computer Vision (ICCV)},
  pages={7589--7599},
  year={2023},
  organization={IEEE Computer Society}
}

@inproceedings{customizeavideo,
  title={Customize-a-video: One-shot motion customization of text-to-video diffusion models},
  author={Ren, Yixuan and Zhou, Yang and Yang, Jimei and Shi, Jing and Liu, Difan and Liu, Feng and Kwon, Mingi and Shrivastava, Abhinav},
  booktitle={European Conference on Computer Vision},
  pages={332--349},
  year={2024},
  organization={Springer}
}

@article{anyv2v,
  title={Anyv2v: A plug-and-play framework for any video-to-video editing tasks},
  author={Ku, Max and Wei, Cong and Ren, Weiming and Yang, Huan and Chen, Wenhu},
  journal={arXiv preprint arXiv:2403.14468},
  year={2024}
}

@article{customvideo,
  title={Customvideo: Customizing text-to-video generation with multiple subjects},
  author={Wang, Zhao and Li, Aoxue and Zhu, Lingting and Guo, Yong and Dou, Qi and Li, Zhenguo},
  journal={arXiv preprint arXiv:2401.09962},
  year={2024}
}

@inproceedings{dynvideo,
  title={Dynvideo-e: Harnessing dynamic nerf for large-scale motion-and view-change human-centric video editing},
  author={Liu, Jia-Wei and Cao, Yan-Pei and Wu, Jay Zhangjie and Mao, Weijia and Gu, Yuchao and Zhao, Rui and Keppo, Jussi and Shan, Ying and Shou, Mike Zheng},
  booktitle={Proceedings of the IEEE/CVF Conference on Computer Vision and Pattern Recognition},
  pages={7664--7674},
  year={2024}
}

@article{customcrafter,
  title={Customcrafter: Customized video generation with preserving motion and concept composition abilities},
  author={Wu, Tao and Zhang, Yong and Wang, Xintao and Zhou, Xianpan and Zheng, Guangcong and Qi, Zhongang and Shan, Ying and Li, Xi},
  journal={arXiv preprint arXiv:2408.13239},
  year={2024}
}

@inproceedings{videoswap,
  title={Videoswap: Customized video subject swapping with interactive semantic point correspondence},
  author={Gu, Yuchao and Zhou, Yipin and Wu, Bichen and Yu, Licheng and Liu, Jia-Wei and Zhao, Rui and Wu, Jay Zhangjie and Zhang, David Junhao and Shou, Mike Zheng and Tang, Kevin},
  booktitle={Proceedings of the IEEE/CVF Conference on Computer Vision and Pattern Recognition},
  pages={7621--7630},
  year={2024}
}

@article{tweediemix,
  title={TweedieMix: Improving Multi-Concept Fusion for Diffusion-based Image/Video Generation},
  author={Kwon, Gihyun and Ye, Jong Chul},
  journal={arXiv preprint arXiv:2410.05591},
  year={2024}
}

@inproceedings{motrans,
  title={Motrans: Customized motion transfer with text-driven video diffusion models},
  author={Li, Xiaomin and Jia, Xu and Wang, Qinghe and Diao, Haiwen and Ge, Mengmeng and Li, Pengxiang and He, You and Lu, Huchuan},
  booktitle={Proceedings of the 32nd ACM International Conference on Multimedia},
  pages={3421--3430},
  year={2024}
}

@article{motionshop,
  title={MotionShop: Zero-Shot Motion Transfer in Video Diffusion Models with Mixture of Score Guidance},
  author={Yesiltepe, Hidir and Meral, Tuna Han Salih and Dunlop, Connor and Yanardag, Pinar},
  journal={arXiv preprint arXiv:2412.05355},
  year={2024}
}

@article{separatemotion,
  title={Separate Motion from Appearance: Customizing Motion via Customizing Text-to-Video Diffusion Models},
  author={Liu, Huijie and Wang, Jingyun and Ma, Shuai and Hu, Jie and Wei, Xiaoming and Kang, Guoliang},
  journal={arXiv preprint arXiv:2501.16714},
  year={2025}
}

@inproceedings{diffusionmotiontransfer,
  title={Space-time diffusion features for zero-shot text-driven motion transfer},
  author={Yatim, Danah and Fridman, Rafail and Bar-Tal, Omer and Kasten, Yoni and Dekel, Tali},
  booktitle={Proceedings of the IEEE/CVF Conference on Computer Vision and Pattern Recognition},
  pages={8466--8476},
  year={2024}
}

@article{diffusion,
  title={Diffusion models beat gans on image synthesis},
  author={Dhariwal, Prafulla and Nichol, Alexander},
  journal={Advances in neural information processing systems},
  volume={34},
  pages={8780--8794},
  year={2021}
}

@inproceedings{lora,
  title={Lora: Low-rank adaptation of large language models},
  author={Hu, Edward J and Shen, Yelong and Wallis, Phillip and Allen-Zhu, Zeyuan and Li, Yuanzhi and Wang, Shean and Wang, Lu and Chen, Weizhu},
  journal={arXiv preprint arXiv:2106.09685},
  year={2021}
}

@article{wan,
  title={Wan: Open and advanced large-scale video generative models},
  author={Wan, Team and Wang, Ang and Ai, Baole and Wen, Bin and Mao, Chaojie and Xie, Chen-Wei and Chen, Di and Yu, Feiwu and Zhao, Haiming and Yang, Jianxiao and others},
  journal={arXiv preprint arXiv:2503.20314},
  year={2025}
}

@article{2024cogvideox,
  title={CogVideoX: Text-to-Video Diffusion Models with An Expert Transformer},
  author={Yang, Zhuoyi and Teng, Jiayan and Zheng, Wendi and Ding, Ming and Huang, Shiyu and Xu, Jiazheng and Yang, Yuanming and Hong, Wenyi and Zhang, Xiaohan and Feng, Guanyu and others},
  journal={arXiv preprint arXiv:2408.06072},
  year={2024}
}

@article{2022cogvideo,
  title={CogVideo: Large-scale Pretraining for Text-to-Video Generation via Transformers},
  author={Hong, Wenyi and Ding, Ming and Zheng, Wendi and Liu, Xinghan and Tang, Jie},
  journal={arXiv preprint arXiv:2205.15868},
  year={2022}
}

@article{hunyuanvideo,
  title={Hunyuanvideo: A systematic framework for large video generative models},
  author={Kong, Weijie and Tian, Qi and Zhang, Zijian and Min, Rox and Dai, Zuozhuo and Zhou, Jin and Xiong, Jiangfeng and Li, Xin and Wu, Bo and Zhang, Jianwei and others},
  journal={arXiv preprint arXiv:2412.03603},
  year={2024}
}

@article{dualreal,
  title={DualReal: Adaptive Joint Training for Lossless Identity-Motion Fusion in Video Customization},
  author={Wang, Wenchuan and Huang, Mengqi and Tu, Yijing and Mao, Zhendong},
  journal={arXiv preprint arXiv:2505.02192},
  year={2025}
}

@article{freelong,
  title={Freelong: Training-free long video generation with spectralblend temporal attention},
  author={Lu, Yu and Liang, Yuanzhi and Zhu, Linchao and Yang, Yi},
  journal={Advances in Neural Information Processing Systems},
  volume={37},
  pages={131434--131455},
  year={2024}
}

@article{riflex,
  title={Riflex: A free lunch for length extrapolation in video diffusion transformers},
  author={Zhao, Min and He, Guande and Chen, Yixiao and Zhu, Hongzhou and Li, Chongxuan and Zhu, Jun},
  journal={arXiv preprint arXiv:2502.15894},
  year={2025}
}

@article{qwen3,
    title={Qwen3 Technical Report}, 
    author={An Yang and Anfeng Li and Baosong Yang and Beichen Zhang and Binyuan Hui and Bo Zheng and Bowen Yu and Chang Gao and Chengen Huang and Chenxu Lv and Chujie Zheng and Dayiheng Liu and Fan Zhou and Fei Huang and Feng Hu and Hao Ge and Haoran Wei and Huan Lin and Jialong Tang and Jian Yang and Jianhong Tu and Jianwei Zhang and Jianxin Yang and Jiaxi Yang and Jing Zhou and Jingren Zhou and Junyang Lin and Kai Dang and Keqin Bao and Kexin Yang and Le Yu and Lianghao Deng and Mei Li and Mingfeng Xue and Mingze Li and Pei Zhang and Peng Wang and Qin Zhu and Rui Men and Ruize Gao and Shixuan Liu and Shuang Luo and Tianhao Li and Tianyi Tang and Wenbiao Yin and Xingzhang Ren and Xinyu Wang and Xinyu Zhang and Xuancheng Ren and Yang Fan and Yang Su and Yichang Zhang and Yinger Zhang and Yu Wan and Yuqiong Liu and Zekun Wang and Zeyu Cui and Zhenru Zhang and Zhipeng Zhou and Zihan Qiu},
    journal = {arXiv preprint arXiv:2505.09388},
    year={2025}
}

@INPROCEEDINGS{clgc,
  author={Xu, Xuancheng and Tao, Ming and Bao, Bing-Kun},
  booktitle={2025 IEEE International Conference on Multimedia and Expo (ICME)}, 
  title={CLGC: Continuous Layout Guidance for Consistent Text-to-Video Editing}, 
  year={2025},
  pages={1-6},
  doi={10.1109/ICME59968.2025.11210198}
}

@inproceedings{det,
  title={Decouple and track: Benchmarking and improving video diffusion transformers for motion transfer},
  author={Shi, Qingyu and Wu, Jianzong and Bai, Jinbin and Zhang, Jiangning and Qi, Lu and Tong, Yunhai and Li, Xiangtai},
  booktitle={Proceedings of the IEEE/CVF International Conference on Computer Vision},
  pages={10995--11005},
  year={2025}
}
